\documentclass[10pt,twocolumn,letterpaper]{article}

\usepackage{iccv}
\usepackage{times}
\usepackage{epsfig}
\usepackage{graphicx}
\usepackage{amsmath}
\usepackage{amssymb}

\usepackage[rightcaption]{sidecap}
\usepackage{subcaption}
\usepackage{multirow}
\usepackage{hhline}
\usepackage[toc,page]{appendix}

\usepackage[pagebackref=true,breaklinks=true,letterpaper=true,colorlinks,bookmarks=false]{hyperref}

\iccvfinalcopy 

\def\iccvPaperID{7804} 

\ificcvfinal\fi

\begin{document}

\title{Patch Craft: Video Denoising by Deep Modeling and Patch Matching}

\author{Gregory Vaksman\\
CS Department - The Technion\\
Technion City, Haifa, Israel\\
{\tt\small grishav@campus.technion.ac.il}
\and
Michael~Elad\\
Google Research\\
Mountain-View, California\\
{\tt\small melad@google.com}
\and
Peyman Milanfar\\
Google Research\\
Mountain-View, California\\
{\tt\small milanfar@google.com}
}

\maketitle

\graphicspath{{./Figures/diagrams/pictures/}{./Figures/diagrams/schemes/}{./Figures/results/}}

\begin{abstract}
The non-local self-similarity property of natural images has been exploited extensively for solving various image processing problems.  When it comes to video sequences, harnessing this force is even more beneficial due to the temporal redundancy. In the context of image and video denoising, many classically-oriented algorithms employ self-similarity, splitting the data into overlapping patches, gathering groups of similar ones and processing these together somehow. With the emergence of convolutional neural networks (CNN), the patch-based framework has been abandoned. Most CNN denoisers operate on the whole image, leveraging non-local relations only implicitly by using a large receptive field. This work proposes a novel approach for leveraging self-similarity in the context of video denoising, while still relying on a regular convolutional architecture. We introduce a concept of \textbf{patch-craft frames} -- artificial frames that are similar to the real ones, built by tiling matched patches. Our algorithm augments video sequences with patch-craft frames and feeds them to a CNN. We demonstrate the substantial boost in denoising performance obtained with the proposed approach. 

\end{abstract}

\section{Introduction}
In this paper we put our emphasis on the denoising task, removing white additive Gaussian noise of known variance from visual content, focusing on video sequences. Image and video denoising is a rich and heavily studied topic, with numerous classically oriented methods and ideas that span decades of extensive research activity. The recent emergence of deep learning has brought a further boost to this field, with better performing solutions. Our goal in this paper to propose a novel video denoising strategy that builds on a synergy between the classics and deep neural networks. A key feature we build upon is the evident spatio-temporal self-similarity existing in video sequences.

Natural images are known to have a spatial \emph{self-similarity} property -- local image components tend to repeat themselves inside the same picture~\cite{zontak2011internal}. Imagine an image split into overlapping patches of small size (e.g.\ $7 \times 7$). Many of these are likely to have several similar twins in different locations in the same image. This property has been exploited extensively by classically oriented algorithms for solving various image processing problems -- denoising and many other tasks. These algorithms usually split the processed image into fully overlapping patches and arrange them into some structure according to their similarity. For example, the well-known Non-Local-Means algorithm~\cite{buades2005non} filters each patch by averaging it with similar ones. The methods reported in~\cite{dabov2007image,mairal2009non} group the similar patches and denoise them jointly. Alternatively, the authors of~\cite{ram2014patch,vaksman2016patch} chain all the patches into a shortest-path, using this as a regularization for solving various inverse problems. Other methods go even farther and construct more complicated structures, such as binary trees~\cite{ram2011generalized} or graphs~\cite{yankelevsky2016dual}, and use these structures for solving image reconstruction tasks. 

In recent years convolutional neural networks (CNN)  entered the image restoration field and took the lead, showing impressive results~(e.g.,\ \cite{zhang2017beyond,zhang2018ffdnet,mao2016image,tai2017memnet,liu2018multi,liu2018non,zhang2020residual}). With this trend in place, the patch-based framework has been nearly abandoned, despite its success and popularity in classical algorithms.  Most CNN based schemes work globally, operating on the whole image rather than splitting it into patches, leveraging self-similarity only implicitly by using a large receptive field. This trend has two origins: First, the reconstructed patches tend to be inconsistent on their overlaps. This undesirable and challenging phenomenon is referred to in the literature as the \emph{local-global} gap, handled typically by plain averaging~\cite{romano2015patch,batenkov2017global,sulam2015expected,zoran2011learning}. The second reason for abandoning the patch-based framework is the difficulty of combining it with CNNs. The convolutional architecture has been shown to be a very successful choice, achieving state-of-the-art (SOTA) results in many image restoration tasks (e.g.,~\cite{zhang2017beyond, mao2016image, tai2017memnet, liu2018multi, zhang2020residual}). However, such an architecture implies working on the whole image uniformly, and thus combining it with a patch-based framework is not straightforward.

Several recent denoisers have combined the patch-based point of view within a deep-learning solution (e.g.~\cite{lefkimmiatis2017non,lefkimmiatis2018universal,vaksman2020lidia}).  Without diving into their details, these algorithms split a noisy image into fully overlapping patches, augment each with a group of similar ones and feed these groups to a denoising network. All three algorithms have managed to achieve near SOTA results while using a small number of trainable parameters. However, their performance is still challenged by leading convolutional networks, such as DnCNN and other networks~\cite{zhang2017beyond,zhang2020residual,liu2018multi,tai2017memnet}.

When turning to video processing, self similarity is further amplified due to the temporal redundancy. Thus, harnessing non-local processing in video is expected to be even more effective and beneficial. Many classical algorithms have successfully exploited spatio-temporal self-similarity by working on 2D or even 3D patches. For example, the V-BM4D~\cite{maggioni2011video} groups similar 3D patches and denoises them by a joint transform and thresholding, extending the well-known BM3D to video~\cite{dabov2007image}. VNLB~\cite{arias2018video} also relies on groups of such patches, employing a joint empirical Bayes estimation for each group, under a Gaussianity modeling. 

In contrast to the activity in image denoising, where many CNN-based schemes surpass classical algorithms, only a few video denoising networks have been shown to be competitive with classical methods. The recently published DVDnet~\cite{tassano2019dvdnet} and FastDVDnet~\cite{tassano2020fastdvdnet} are such networks, obtaining SOTA results, the first leveraging motion compensation, and the second combining several U-Net~\cite{ronneberger2015u} networks for increasing its receptive field. Other CNN-based video denoisers in recent literature are~\cite{yue2020supervised,claus2019videnn,xue2019video,ehret2019model}

While CNN-based algorithms for video denoising, such as DVDnet and FastDVDnet, choose to work on whole frames rather than on patches, neural network based video denoising may still exploit self-similarity. For example, VNLnet~\cite{davy2019non}, the first good-performing such denoiser, combines the patch-based framework with DnCNN~\cite{zhang2017beyond} architecture, augmenting noisy frames with auxiliary feature maps that consist of central pixels taken from similar patches. While VNLnet's strategy introduces a non-locality flavor into the denoising process, it is limited in its effectiveness due to the use of central pixels instead of full neighboring patches, as evident from its performance.

Our work proposes a novel, intuitive, and highly effective way to leverage non-local self-similarity within a CNN architecture. We introduce the concept of \emph{patch-craft} frames and use these as feature maps within the denoising process. For constructing the patch-craft frames, we split each video frame into fully overlapping patches. For each patch we find its $n$ nearest neighbors in a spatio-temporal window, and those are used to build $f$ (patch size) groups of corresponding $n$ patch-craft frames. These are augmented to the real video frames and fed into a spatio-temporal denoising network. This way, self-similarity is fully leveraged, while preserving the CNN's nature of operating on whole frames, and overcoming the local-global gap.

Augmenting video sequences with patch-craft frames requires processing large amounts of data in producing each output frame.  To overcome this difficulty, we use a CNN composed of multidimensional \emph{separable convolutional} (SepConv) layers. A SepConv layer applies a series of convolutional filters, each working on a sub-group of dimensions while referring to the rest as independent tensors. Such layers implement a multidimensional separable convolution, which allows reducing the number of trainable parameters and expediting the inference. 

The processing pipeline of our proposed augmentation scheme for video denoising is composed of two stages. First, we augment a noisy video sequence with patch-craft frames and feed the augmented sequence into a CNN built of SepConv layers. This stage functions mostly as a spatial filtering. At the second stage, we apply a temporal filtering, using a 3D extension of the DnCNN~\cite{zhang2017beyond} architecture. This filter works in a sliding window manner, getting as input the outcome of the first stage with the original noisy video and producing reconstructed video at its output. Through extensive experiments, we show that the proposed method leads to a substantial boost in video denoising performance compared with leading SOTA algorithms.

To summarise, the contributions of this work are the following: We propose a neural network based video denoising scheme that consists of an augmentation of patch-craft frames, followed by a spatial and a temporal filtering. The proposed augmentation leverages non-local self-similarity using the patch-based framework, while allowing the denoising network to operate on whole frames. The deployed SepConv layers, which are used as building blocks of the spatial filtering CNN, allow reasonable memory and computational complexities for inference and learning, despite the large number of the patch-craft frames. The proposed method shows SOTA video denoising performance when compared to leading alternative algorithms.


\section{Patch-Craft Frames} \label{sec:patch_craft_frames}
Let us start by motivating our approach. Consider a video frame that should be denoised by a neural network. Imagine that one could construct an artificial frame identical to the real one but with a different noise realization. Such a synthetic frame would be beneficial for denoising because it holds additional information about the processed frame. More specifically, this synthetic frame could be used as an additional feature map representing the real frame. Following this motivation, we define the patch-craft frames as such auxiliary artificial frames built of patches taken from the current and surrounding frames.

\begin{SCfigure}[1.8]
    \caption{Patches of size ${\sqrt{F} \times \sqrt{F}}$ are used for nearest neighbor search. Their central ${\sqrt{f} \times \sqrt{f}}$ part are used for constructing patch-craft frames.}
	\label{fig:nn_dn_patch}
    \includegraphics[width=0.15\textwidth]{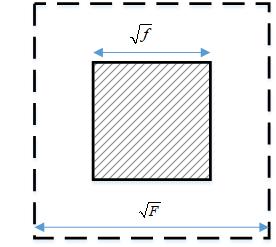}
\end{SCfigure}

Constructing of the patch-craft frames is carried out as follows: We start by extracting all possible overlapping patches of size $\sqrt{F} \times \sqrt{F}$ from the processed frame, which we refer to as the \emph{current} frame, and find $n$ nearest neighbors (most similar patches) for each extracted patch. We use the $L_2$ norm as distance metrics and limit the nearest neighbor search to a spatio-temporal $3D$ box of size ${B \times B \times \left(2T_s + 1\right)}$, where B refers to spatial axes and $2T_s + 1$ stands for the temporal window used - $T_s$ backward and $T_s$ forward. The $n$ found neighbor patches are used for building the patch-craft frames where we utilize only their central parts of size $\sqrt{f} \times \sqrt{f}$, as shown in Figure~\ref{fig:nn_dn_patch}. 
    
\begin{figure}
    \centering
	\begin{subfigure}{0.23\textwidth}
	    \captionsetup{justification=centering}
		\includegraphics[width=\textwidth]{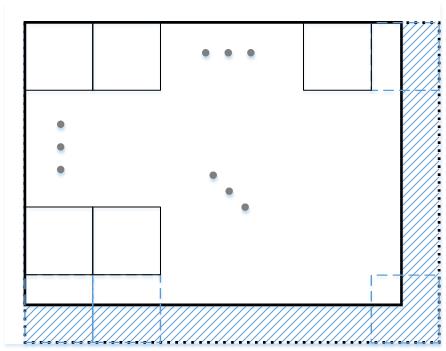}
		\caption{Offset $\left[0, 0\right]$}
		\label{fig:patch-craft:no_offs}
	\end{subfigure}
	\begin{subfigure}{0.23\textwidth}
	    \captionsetup{justification=centering}
		\includegraphics[width=\textwidth]{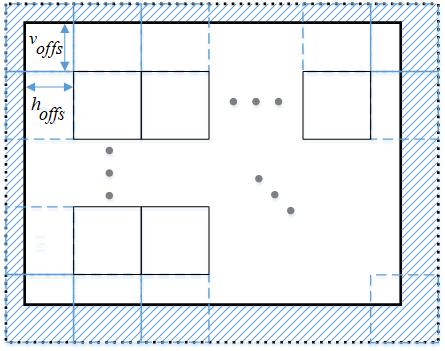}
		\caption{Offset $\left[v_{offs}, h_{offs}\right]$}
		\label{fig:patch-craft:offs}
	\end{subfigure}
	\caption{Splitting a frame to non-overlapping patches with different offsets. Figure~\ref{fig:patch-craft:no_offs} shows a splitting to patches without an offset ($[0, 0]$), while~\ref{fig:patch-craft:offs} shows a splitting to patches with an offset $\left[v_{offs}, h_{offs}\right]$. The white rectangle represents the processed frame, where the blue area  represents a mirror reflection of the frame pixels.} 
	\label{fig:patch-craft}
\end{figure}

The patch-craft frames are created by stitching non-overlapping patches together. More specifically, we build $f$ groups of $n + 1$ frames. Each group contains a copy of the processed frame and $n$ patch-craft ones. The first patch-craft frame is built by stitching the first nearest neighbors together, the second frame is constructed from the second nearest neighbors, and so on, till the last ($n\text{-th}$) nearest neighbor. The $f$ groups differ by the patches' offsets.  For building the first group, we use the neighbors of patches with no offset (i. e., with offset $[0, 0]$). The second group is constructed using the neighbors of patches with offset $[0, 1]$, and so on, till an offset $[\sqrt{f} - 1, \sqrt{f} - 1]$. For handling boundary pixels, we extrapolate the frame with a mirror reflection of itself and cut the leftovers after stitching the neighbors. Splitting of a frame to non-overlapping patches with different offsets is shown schematically in Figure~\ref{fig:patch-craft}. 

\begin{figure*}
    \centering
	\begin{subfigure}{0.23\textwidth}
	    \captionsetup{justification=centering}
		\includegraphics[width=\textwidth]{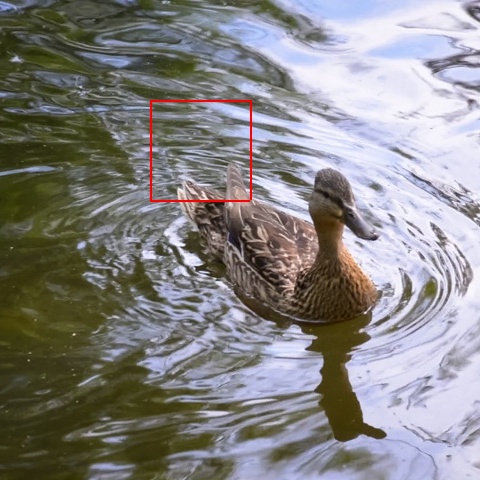}
		\caption{Clean frame}
		\label{fig:blockiness:clean_0_rect}
	\end{subfigure}
	\begin{subfigure}{0.23\textwidth}
	    \captionsetup{justification=centering}
		\includegraphics[width=\textwidth]{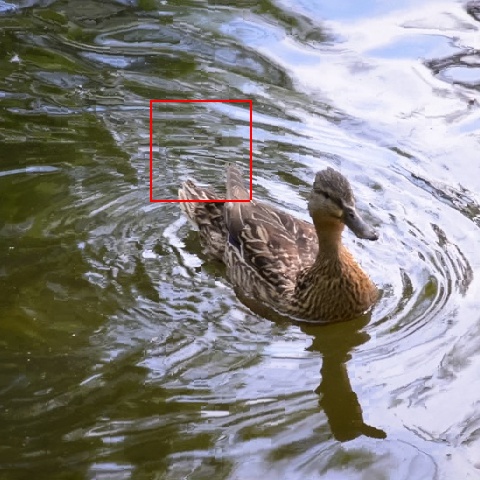}
		\caption{Clean, fifth neighbor}
		\label{fig:blockiness:clean_n_rect}
	\end{subfigure}
	\begin{subfigure}{0.23\textwidth}
	    \captionsetup{justification=centering}
		\includegraphics[width=\textwidth]{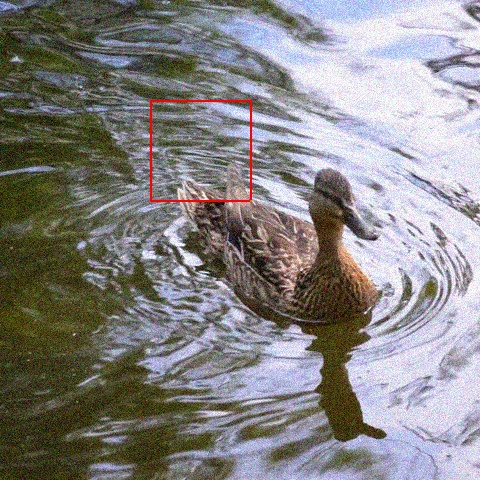}
		\caption{Noisy frame with $\sigma = 25$}
		\label{fig:blockiness:noisy_0_rect}
	\end{subfigure}
	\begin{subfigure}{0.23\textwidth}
	    \captionsetup{justification=centering}
		\includegraphics[width=\textwidth]{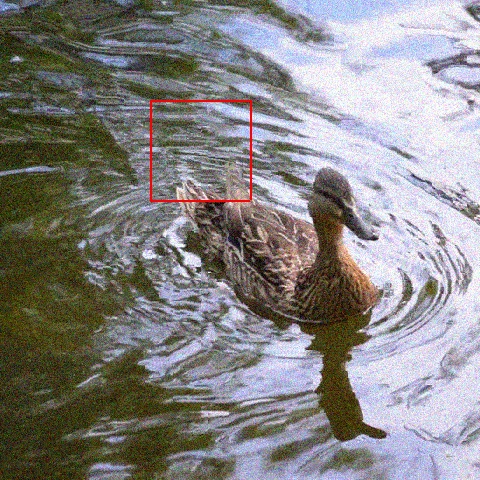}
		\caption{Noisy, fifth neighbor}
		\label{fig:blockiness:noisy_n_rect}
	\end{subfigure}
	\begin{subfigure}{0.23\textwidth}
	    \captionsetup{justification=centering}
		\includegraphics[width=\textwidth]{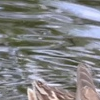}
		\caption{Clean frame}
		\label{fig:blockiness:clean_0_zoom}
	\end{subfigure}
	\begin{subfigure}{0.23\textwidth}
	    \captionsetup{justification=centering}
		\includegraphics[width=\textwidth]{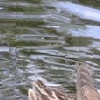}
		\caption{Clean, fifth neighbor}
		\label{fig:blockiness:clean_n_zoom}
	\end{subfigure}
	\begin{subfigure}{0.23\textwidth}
	    \captionsetup{justification=centering}
		\includegraphics[width=\textwidth]{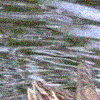}
		\caption{Noisy frame with $\sigma = 25$}
		\label{fig:blockiness:noisy_0_zoom}
	\end{subfigure}
	\begin{subfigure}{0.23\textwidth}
	    \captionsetup{justification=centering}
		\includegraphics[width=\textwidth]{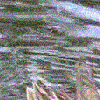}
		\caption{Noisy, fifth neighbor}
		\label{fig:blockiness:noisy_n_zoom}
	\end{subfigure}
	\caption{An example of the block boundary artifacts and the influence of the additive random Gaussian noise. This figure shows the fifth patch-craft frame, i.e., built of fifth neighbors, for frame 6 of the sequence \emph{mallard-water}. As can be seen by comparing~\ref{fig:blockiness:clean_n_rect},~\ref{fig:blockiness:clean_n_zoom} and~\ref{fig:blockiness:noisy_n_rect},~\ref{fig:blockiness:noisy_n_zoom}, the patch-craft frame built using a clean sequence suffers from block boundary artifacts while the noisy data leads to artifact reduction.} 
	\label{fig:blockiness}
\end{figure*}

Clearly, stitching patches with no overlaps to form an image may lead to the block boundary artifacts. Such artifacts appear, for example, in heavy JPEG compression. Therefore, the reader may wonder how do we avoid these artifacts in the patch-craft frames? Indeed, a naive attempt to construct patch-craft frames from a clean video sequence may lead to a significant block boundary artifacts. These are clearly seen in Figures~\ref{fig:blockiness:clean_n_rect}~and~\ref{fig:blockiness:clean_n_zoom}, which show an example of stitching together clean fifth nearest neighbors. 

In order to explain why this problem is avoided in our case, we draw intuition from dithering methods. It is well-known that adding random noise before quantization causes a reduction in visual artifacts. More generally, adding random noise to a signal can help to combat structural noise. An adaptation of this idea for our case is immediate: the fact that the handled video sequence is already noisy leads to reduced artifacts, as can be seen in an example in Figure~\ref{fig:blockiness}.  A comparison between Figures~\ref{fig:blockiness:clean_n_rect},~\ref{fig:blockiness:clean_n_zoom} and Figures~\ref{fig:blockiness:noisy_n_rect},~\ref{fig:blockiness:noisy_n_zoom} exposes the benefit of the added random noise. 

In addition to the $f$ groups of $n + 1$ frames, we provide the denoising network with $n + 1$ feature maps of scores that indicate the reliability of the patch-craft frames. A natural measure for this reliability is a patchwise squared distance between the processed and the patch-craft frames.
If we denote the processed frame by $y$, and the $j$th patch-craft frame in group $i$ by $\tilde{y}_{ij}$, then the patchwise squared distance $d_{ij}$ can be calculated by subtracting the frames, computing the pointwise square of the difference, and convolving the result with a uniform kernel,
\begin{equation}
    d_{ij} = \operatorname{conv2d}\left(\left(y - \tilde{y}_{ij}\right) ** 2, \operatorname{ones}\left(\sqrt{f}, \sqrt{f}\right)\right) \;.
\end{equation}
Since the neural network can learn and absorb convolution kernels, we omit the last convolution. Thus we build the feature maps of scores by calculating average pointwise squared distances between the processed and patch-craft frames. More specifically, these feature maps is a group of ${n + 1}$ frames $\left\{d_j\right\}_{j = 0}^n$\footnote{We set $\tilde{y}_{0j} = y$, thus $d_0$ is zero, used for preserving tensor size.}, where
\begin{equation}
    d_j = \frac{1}{f} \sum_{i = 0}^{f - 1} \left(y - \tilde{y}_{ij}\right) ** 2 \;.
\end{equation}
We concatenate these feature maps of scores with the $f$ groups of $n + 1$ frames along the $f$ dimension, passing the denoising network $f + 1$ groups of $n + 1$ frames for each processed original frame.


\section{The Proposed Algorithm}
In this section we present the proposed architecture covering the separable spatial and the temporal filters.

\subsection{Separable Convolutional Neural Network}
As described in the previous section, the spatial denoising network gets as input ${f + 1}$ groups of ${n + 1}$ feature maps for each processed frame. For instance, if the patch size is $7 \times 7$ (i.e., $f = 49$) and $n = 14$, the number of feature maps pushed into the network for each processed frame (3 color channels) is ${\left(49 + 1\right) \times \left(14 + 1\right) \times 3 = 2250}$. A regular convolutional neural network would be inapplicable for processing such an amount of data, as even using the smallest possible filters of size $3 \times 3$ requires learning kernels of size $3 \times 3 \times 2250$. To overcome this difficulty, we use separable convolutional layers.
\begin{figure}
    \centering
	\begin{subfigure}{0.46\textwidth}
	    \captionsetup{justification=centering}
		\includegraphics[width=\textwidth]{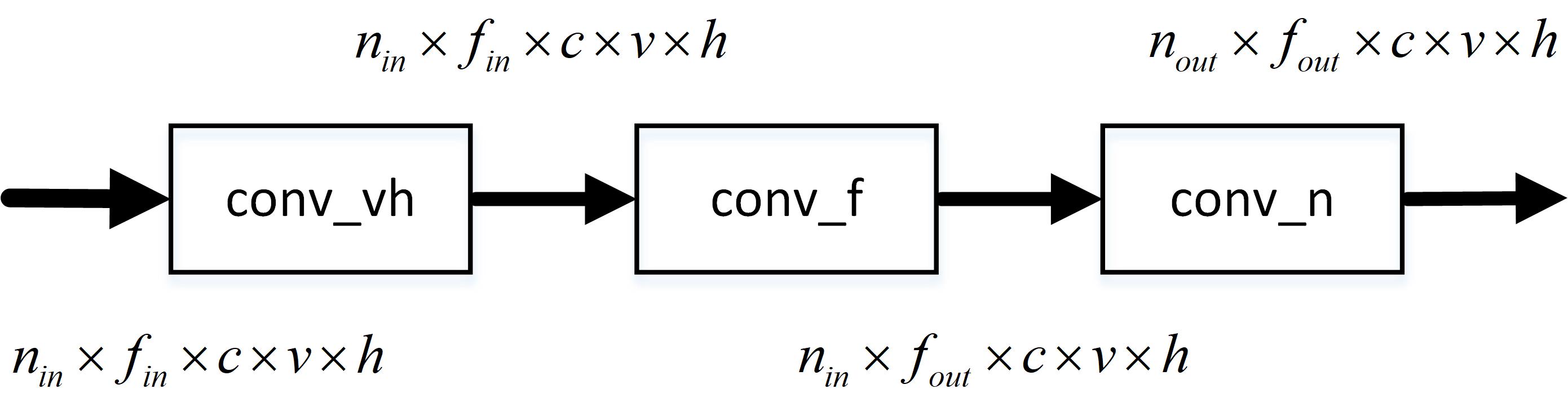}
	\end{subfigure}
	\caption{The SepConv layer.}
	\label{fig:sep_conv}
\end{figure}

The proposed SepConv layer is shown schematically in Figure~\ref{fig:sep_conv}. It is a separable convolutional layer composed of three convolutional filters, $conv\_vh$, $conv\_f$, and $conv\_n$. Each filter works on a sub-group of dimensions and refers to the rest as independent tensors. The input and output of SepConv are five dimensional tensors of sizes ${n_{in} \times f_{in} \times c \times v \times h}$ and ${n_{out} \times f_{out} \times c \times v \times h}$ respectively, where $[v,h]$ is the frame size, $c$ is the number of color layers, $f$ is the patch-size, and $n$ is the number of neighbors to be used.

The $conv\_vh$ filter applies $2D$ convolutions with kernels of size $m \times m$ referring to dimension $c$ as input channels and dimensions $n$ and $f$ as independent ones. This filter represents a local spatial prior, having $n_{in}f_{in}$ groups of trainable convolution kernels. The $conv\_f$ filter applies $2D$ convolutions with $1 \times 1$ kernels referring to dimensions $c$ and $f$ as input channels and $n$ as independent. This filter represents a weighted patch averaging, having $n_{in}$ groups of trainable kernels.  $conv\_n$ applies $2D$ convolutions with $1 \times 1$ kernels referring to $n$ as input channels, while $c$ and $f$ are referred to as independent. This kernel represents a weighted neighbor averaging, having $f_{out}c$ groups of trainable kernels.

\begin{figure}
    \centering
	\begin{subfigure}{0.4\textwidth}
	    \captionsetup{justification=centering}
		\includegraphics[width=\textwidth]{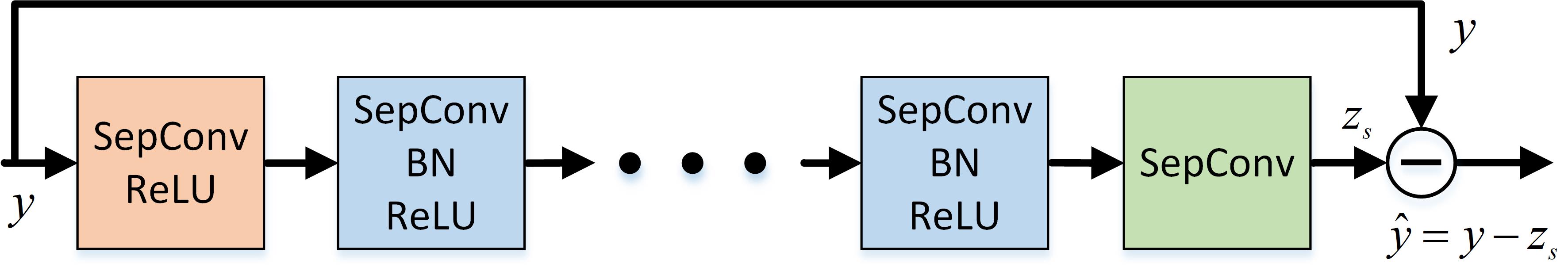}
	\end{subfigure}
	\caption{Our spatial denoising network S-CNN.}
	\label{fig:spatial_cnn}
\end{figure}

The spatial denoising network (S-CNN) is composed of blocks as shown in Figure~\ref{fig:spatial_cnn}. The first block includes the SepConv layer followed by ReLU, the middle blocks are similar to the first but with a Batch Normalization (BN) between SepConv and ReLU, and the last block consists of a single SepConv layer. Each SepConv layer reduces the number of neighbors by a factor of 2, i.e., ${n_{out} = \left \lceil n_{in} / 2 \right \rceil}$. The network operates in the residual domain predicting the noise $z_s$. The output frame $\hat{y}$ is obtained by subtracting the predicted noise from the corrupted frame $y$.


\subsection{Temporal Filtering}
\begin{figure}
    \centering
    \begin{subfigure}[b]{.18\textwidth}
        \includegraphics[width=\textwidth]{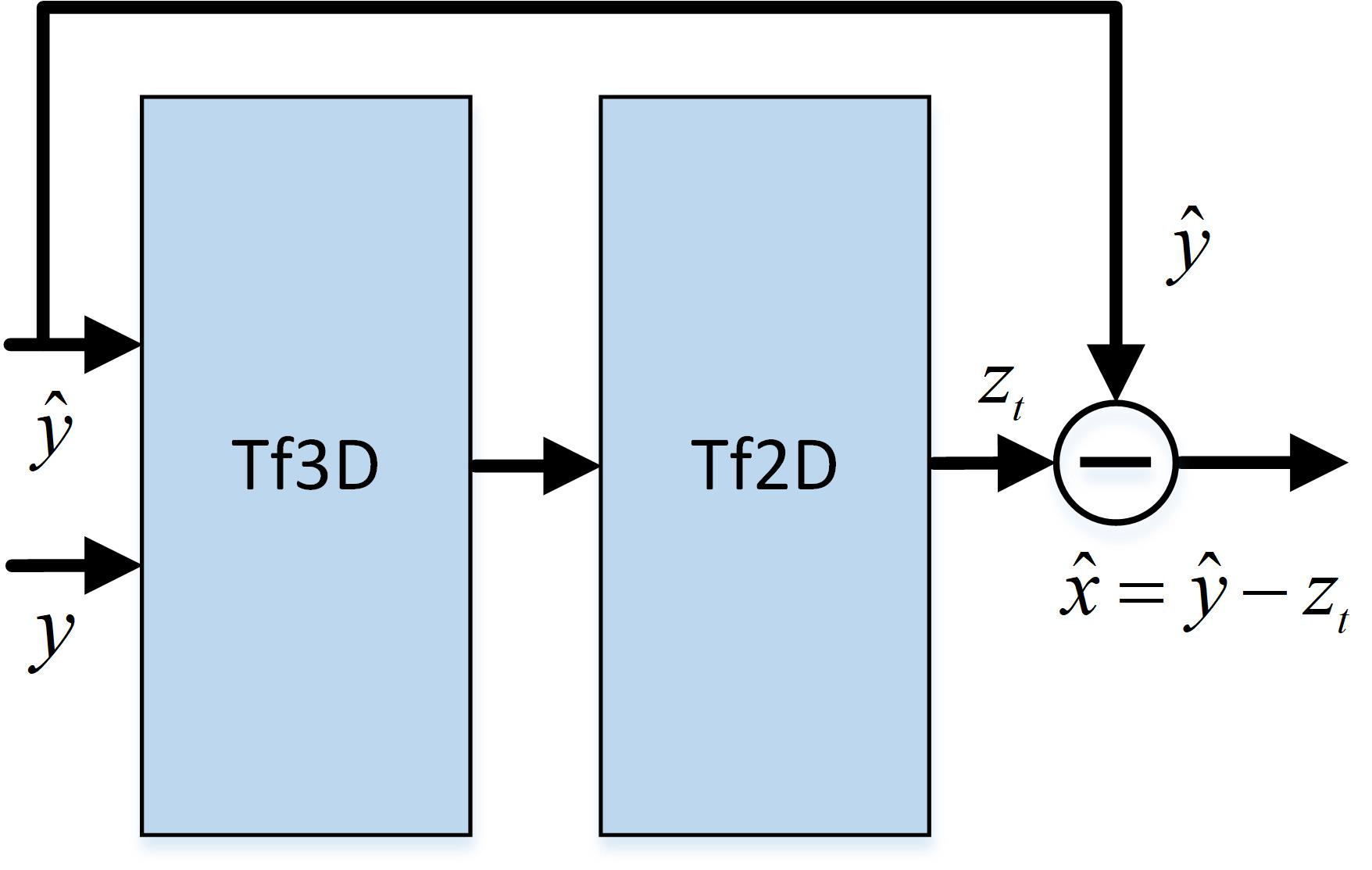}
        \caption{Temporal denoising network T-CNN}
	\label{fig:temporal:tf_cnn}
    \end{subfigure}
    \quad
    \begin{subfigure}[b]{.18\textwidth}
        \includegraphics[width=\textwidth]{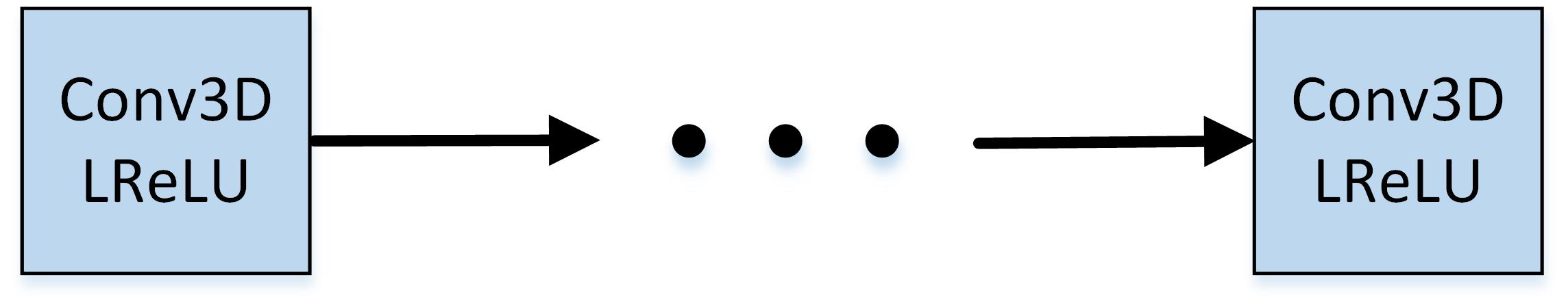}
        \caption{Tf3D}
    	\label{fig:temporal:tf_3d}
        
        \vspace{2ex}
        
        \includegraphics[width=\textwidth]{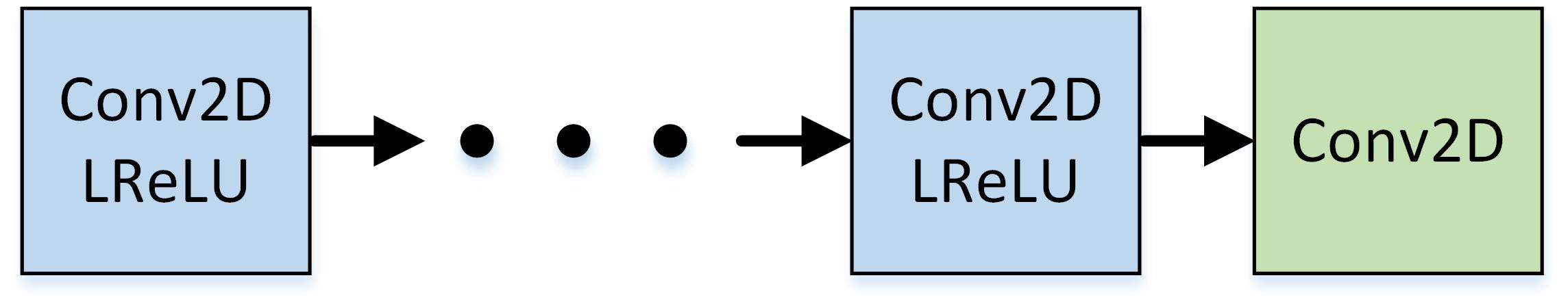}
        \caption{Tf2D}
    	\label{fig:temporal:tf_2d}
    \end{subfigure}
    \caption{Our temporal filtering network.}
	\label{fig:temporal}
\end{figure}

Although S-CNN processes information from adjacent frames due to the augmentation, it does not guarantee temporal continuity. More generally, S-CNN does not impose an explicit temporal prior on the denoised video sequence. Thus, we apply temporal post-filtering, T-CNN, on the S-CNN output. The architecture of T-CNN is shown in Figure~\ref{fig:temporal}, working in a sliding window -- getting as input ${2T_t + 1}$ frames for any output one. Each T-CNN input frame is a concatenation (along the color dimension) of the S-CNN input and output frames $y$ and $\hat{y}$. Similar to S-CNN, T-CNN works in the residual domain, predicting the noise $z_t$. The output frame $\hat{x}$ is obtained by subtracting the predicted noise from the partially denoised frame $\hat{y}$. The network architecture should remind the reader of DnCNN~\cite{zhang2017beyond}. The first part of it, Temporal Filter 3D (Tf3D), is composed of $T_t$ blocks consisting of a $3D$ convolutions with ${3 \times 3 \times 3}$ kernels followed by Leaky ReLUs (LReLU). The second part, Temporal Filter 2D (Tf2D), consists of $2D$ convolutions with ${3 \times 3}$ kernels followed by LReLU. Each 3D kernel of Tf3D applies no padding in the temporal dimension, while padding with zeros spatially. The kernels of Tf2D apply padding with zeros as well.




\section{Experimental Results}

\subsection{Video Denoising}
In this section we report the denoising performance of our scheme, while comparing it with leading algorithms. We refer hereafter to our method as Patch-Craft Network (PaCNet). For quantitative comparisons, we use the PSNR metric, which is a commonly used measure of distortion for reconstructed video. In addition, we present qualitative comparisons between the reconstructed videos. Among classical video denoisers, we compare with V-BM4D~\cite{maggioni2011video} and VNLB~\cite{arias2018video} as they are the best performing classical schemes in terms of PSNR. In comparisons with CNN-based denoisers, we include VNLnet~\cite{davy2019non} since it has a flavor of non-locality, and the two SOTA networks, DVDnet~\cite{tassano2019dvdnet} and FastDVDnet~\cite{tassano2020fastdvdnet}.

\begin{table*}
    \centering
    \setlength{\doublerulesep}{1pt}
    \begin{tabular}{|c||c|c|c|c|c|c|c|c|}
        \hline
        \multirow{3}{*}{Noise $\sigma$} & \multicolumn{8}{c|}{Method} \\
        \hhline{|~||--------|}
        & $\operatorname{V-BM4D}$ & $\operatorname{VNLB}$ & $\operatorname{VNLnet}$ & $\operatorname{DVDnet}$ & $\operatorname{FastDVDnet}$\footnote & $\operatorname{S-CNN-0}$ & $\operatorname{S-CNN-3}$ & $\operatorname{PaCNet}$ \\
        
        & \cite{maggioni2011video} & \cite{arias2018video} & \cite{davy2019non} & \cite{tassano2019dvdnet} & \cite{tassano2020fastdvdnet} & (single frame) & (no $\operatorname{T-CNN}$) & (ours) \\

        \hhline{|=#=|=|=|=|=|=|=|=|}
        $10$ & 37.58 & 38.85 & 35.83 & 38.13 & 38.93 & 38.38 & 39.90 & \textcolor{red}{39.97} \\
        \hline
        $20$ & 33.88 & 35.68 & 34.49 & 35.70 & 35.88 & 34.85 & 36.48 & \textcolor{red}{36.82} \\
        \hline
        $30$ & 31.65 & 33.73 &   -\footnote   & 34.08 & 34.12 & 32.86 & 34.34 & \textcolor{red}{34.79} \\
        \hline
        $40$ & 30.05 & 32.32 & 32.32 & 32.86 & 32.87 & 31.56 & 32.78 & \textcolor{red}{33.34} \\
        \hline
        $50$ & 28.80 & 31.13 & 31.43 & 31.85 & 31.90 & 30.51 & 31.55 & \textcolor{red}{32.20} \\
        \hhline{|=#=|=|=|=|=|=|=|=|}
        Average & 32.39 & 34.34 & - & 34.52 & 34.74 & 33.63 & 35.01 & \textcolor{red}{35.42} \\
        \hline
    \end{tabular}
    \caption{Video denoising performance on DAVIS~\cite{pont20172017} test set: Best PSNR is marked in \textcolor{red}{red}.}
    \label{tab:denoising_known_sig}
\end{table*}
\begin{table}
    \centering
    \setlength{\doublerulesep}{1pt}
    \begin{tabular}{|c||c|c|c||c|}
        \hline
        \multirow{2}{*}{Method} & \multicolumn{3}{c||}{Noise $\sigma$} & \multirow{2}{*}{Average} \\
        \hhline{|~||---||~}
        & 10 & 30 & 50 & \\
        \hhline{|=#=|=|=#=|}
        $\operatorname{ViDeNN}$~\cite{claus2019videnn} & 37.13 &  32.24 & 29.77 & 33.05 \\
        \hline
        $\operatorname{FastDVDnet}$\footnote[2] & 38.65 & 33.59 & 31.28 & 34.51 \\
        \hline
        $\operatorname{PaCNet}$ (ours) & \textcolor{red}{39.96} & \textcolor{red}{34.66} & \textcolor{red}{32.00} & \textcolor{red}{35.54} \\
        \hline
    \end{tabular}
    \caption{Denoising for clipped Gaussian noise.}
    \label{tab:denoising_clipped}
\end{table}
\begin{table}
    \centering
    \setlength{\doublerulesep}{1pt}
    \begin{tabular}{|c||c|c|c||c|}
        \hline
        \multirow{2}{*}{Method} & \multicolumn{3}{c||}{Noise $\sigma$} & \multirow{2}{*}{Average} \\
        \hhline{|~||---||~}
        & 15 & 25 & 50 & \\
        \hhline{|=#=|=|=#=|}
        $\operatorname{LIDIA}$~\cite{vaksman2020lidia} & 34.03 & 31.31 & 27.99 & 31.11 \\
        \hline
        $\operatorname{S-CNN-0}$ (ours) & 33.95 & 31.22 & 27.93 & 31.03 \\
        \hline
    \end{tabular}
    \caption{Single image denoising performance comparison.}
    \label{tab:image_denoising}
\end{table}

We test our network with additive white Gaussian noise of known variance, a standard and common scenario. The algorithm parameters are as follows: ${T_s = 3}$, ${B = 89}$, ${\sqrt{F} = 15}$, ${\sqrt{f} = 7}$, ${n = 14}$. S-CNN has 5 blocks (3 inner SepConv + BN + ReLU blocks) with ${m = 7}$. For the first SepConv layer ${n_{in} = n + 1}$, and ${f_{in} = f + 1}$. For all layers except the last ${f_{out} = f_{in}}$, while for the last ${f_{out} = 1}$. \hbox{T-CNN} has 17 Conv2D layers with 96 channels each, Conv3D layers have 48 channels, and ${T_t = 3}$. Our network is trained on 90 video sequences at 480p resolution -- the DAVIS dataset~\cite{pont20172017}. The spatial and the temporal CNNs are trained separately, both using the Mean Squared Error (MSE) loss. We start by training the spatial CNN alone, and then fix its parameters and train the temporal CNN. Both networks are trained using Lamb optimizer~\cite{you2019large} with a decreasing learning rate, starting from $5 \cdot 10^{-3}$  for the spatial and $2 \cdot 10^{-3}$ for the temporal CNNs. Our network has in total $2.87 \cdot 10^6$ trainable parameters, where $1.34 \cdot 10^6$ are S-CNN parameters and $1.53 \cdot 10^6$ are T-CNN parameters. The inference time for video resolution of $854 \times 480$ pixels is about $0.5$ minute per frame on Nvidia Quadro RTX 8000 GPU or about $5.5$ minutes per frame on CPU.

Table~\ref{tab:denoising_known_sig} reports the average PSNR performance per noise level for 30 test video sequences from the DAVIS dataset (Test-Dev 2017) at 480p resolution. As can be seen, PaCNet shows a substantial boost in denoising performance of 0.5~dB - 1.2~dB, compared with the existing SOTA algorithms. When compared to FastDVDnet and DVDnet, the PSNR benefit decreases with the increase in noise level. This behavior can be explained by the deterioration of the nearest neighbor search for higher noise levels.

\begin{figure*}
    \centering
	\begin{subfigure}{0.32\textwidth}
	    \captionsetup{justification=centering}
		\includegraphics[width=\textwidth]{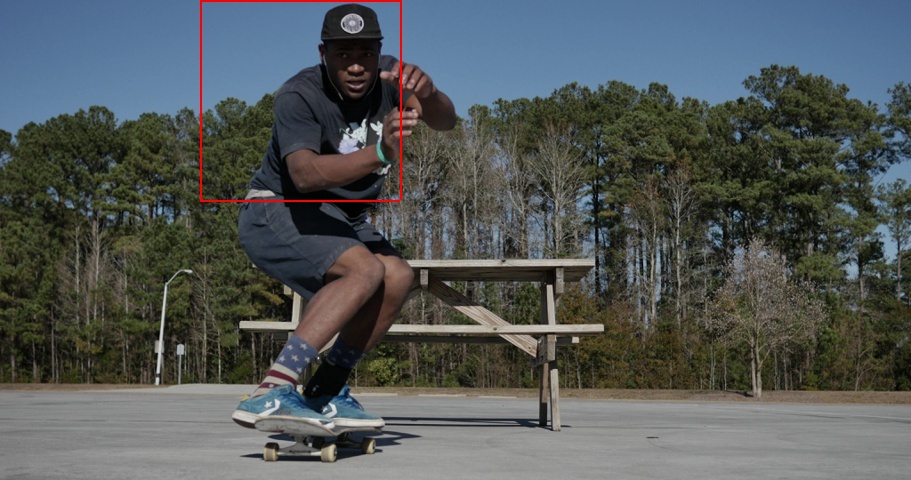}
		\caption{Original}
		\label{fig:skate-jump_f60_s20_c_rect}
		\vspace*{6pt}
	\end{subfigure}
	\begin{subfigure}{0.32\textwidth}
	    \captionsetup{justification=centering}
		\includegraphics[width=\textwidth]{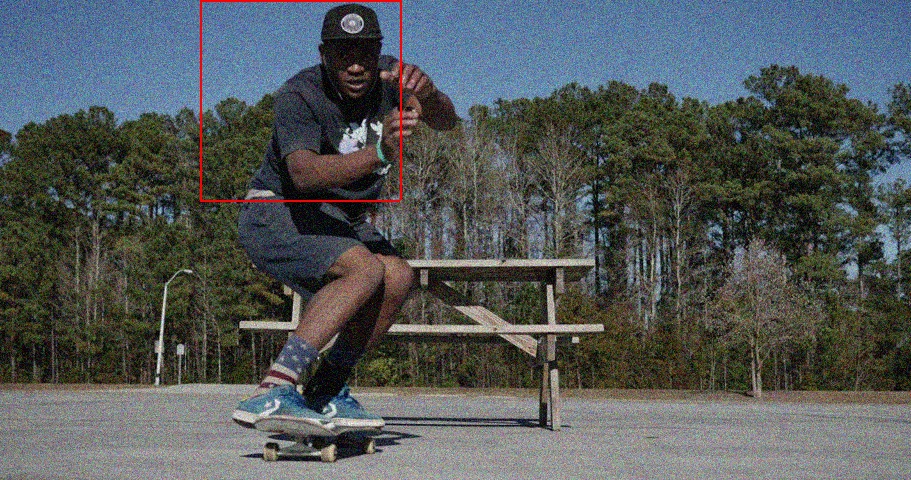}
		\caption{Noisy with $\sigma = 20$}
		\label{fig:skate-jump_f60_s20_n_rect}
		\vspace*{6pt}
	\end{subfigure}
	\begin{subfigure}{0.32\textwidth}
	    \captionsetup{justification=centering}
		\includegraphics[width=\textwidth]{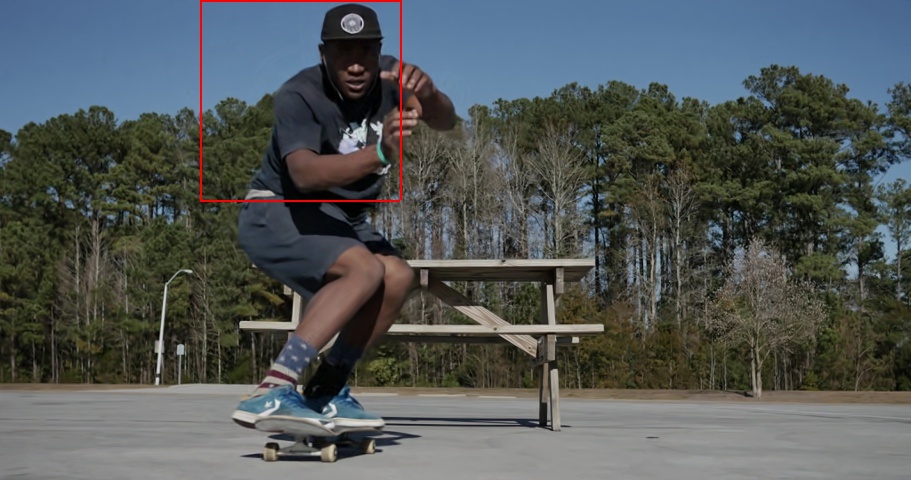}
		\caption{VNLB~\cite{arias2018video}, PSNR = 36.39dB}
		\label{fig:skate-jump_f60_s20_vnlb_rect}
		\vspace*{6pt}
	\end{subfigure}
	\begin{subfigure}{0.32\textwidth}
	    \captionsetup{justification=centering}
		\includegraphics[width=\textwidth]{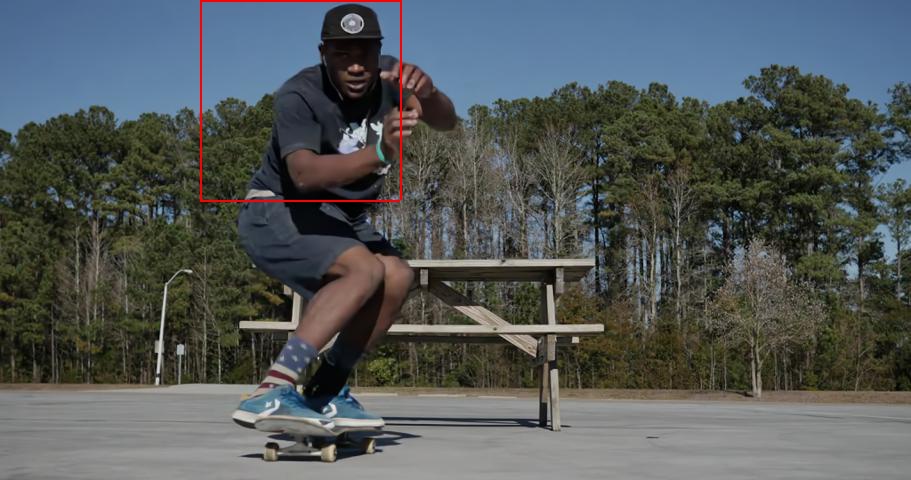}
		\caption{VNLnet~\cite{davy2019non}, PSNR = 33.41dB}
		\label{fig:skate-jump_f60_s20_vnlnet_rect}
		\vspace*{6pt}
	\end{subfigure}
	\begin{subfigure}{0.32\textwidth}
	    \captionsetup{justification=centering}
		\includegraphics[width=\textwidth]{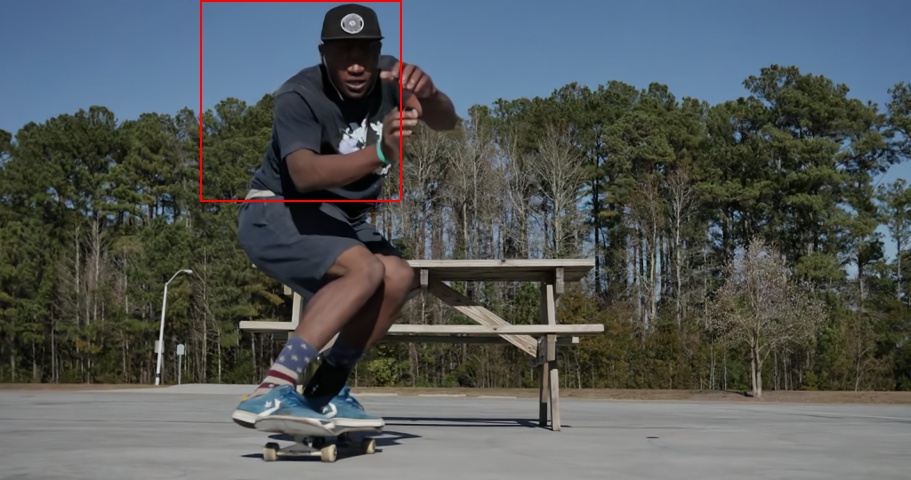}
		\caption{FastDVDnet~\cite{tassano2020fastdvdnet}, PSNR = 35.15dB}
		\label{fig:skate-jump_f60_s20_fastdvd_rect}
		\vspace*{6pt}
	\end{subfigure}
	\begin{subfigure}{0.32\textwidth}
	    \captionsetup{justification=centering}
		\includegraphics[width=\textwidth]{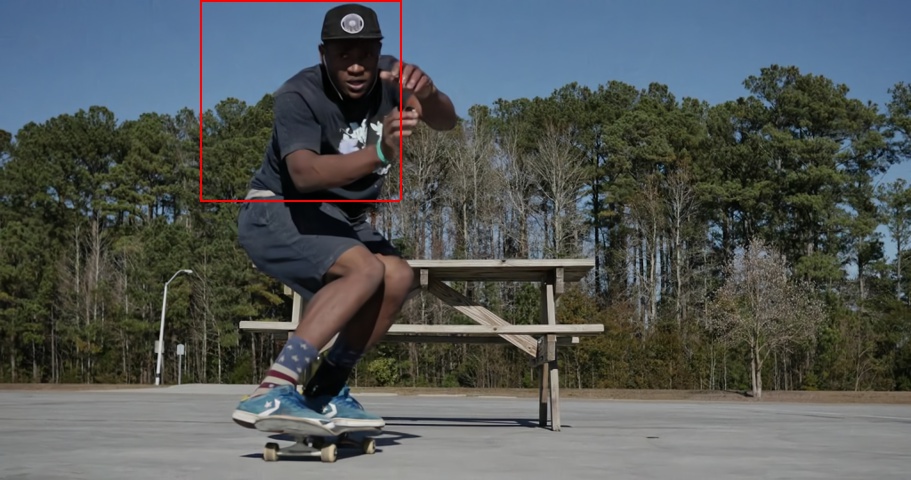}
		\caption{PaCNet (ours), PSNR = 37.31dB}
		\label{fig:skate-jump_f60_s20_covid_rect}
		\vspace*{6pt}
	\end{subfigure}
	\begin{subfigure}{0.32\textwidth}
	    \captionsetup{justification=centering}
		\includegraphics[width=\textwidth]{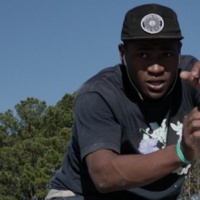}
		\caption{Original}
		\label{fig:skate-jump_f60_s20_c_zoom}
		\vspace*{6pt}
	\end{subfigure}
	\begin{subfigure}{0.32\textwidth}
	    \captionsetup{justification=centering}
		\includegraphics[width=\textwidth]{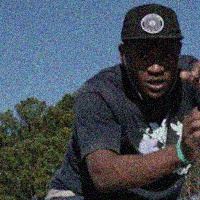}
		\caption{Noisy with $\sigma = 20$}
		\label{fig:skate-jump_f60_s20_n_zoom}
		\vspace*{6pt}
	\end{subfigure}
	\begin{subfigure}{0.32\textwidth}
	    \captionsetup{justification=centering}
		\includegraphics[width=\textwidth]{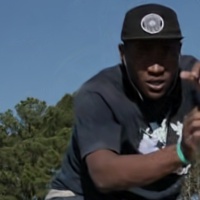}
		\caption{VNLB~\cite{arias2018video}, PSNR = 35.21dB}
		\label{fig:skate-jump_f60_s20_vnlb_zoom}
		\vspace*{6pt}
	\end{subfigure}
	\begin{subfigure}{0.32\textwidth}
	    \captionsetup{justification=centering}
		\includegraphics[width=\textwidth]{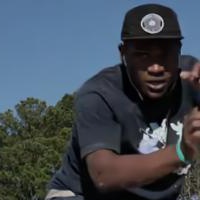}
		\caption{VNLnet~\cite{davy2019non}, PSNR = 33.61dB}
		\label{fig:skate-jump_f60_s20_vnlnet_zoom}
		\vspace*{6pt}
	\end{subfigure}
	\begin{subfigure}{0.32\textwidth}
	    \captionsetup{justification=centering}
		\includegraphics[width=\textwidth]{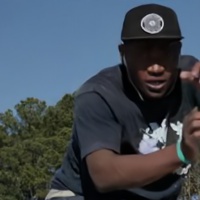}
		\caption{FastDVDnet~\cite{tassano2020fastdvdnet}, PSNR = 35.00dB}
		\label{fig:skate-jump_f60_s20_fastdvd_zoom}
		\vspace*{6pt}
	\end{subfigure}
	\begin{subfigure}{0.32\textwidth}
	    \captionsetup{justification=centering}
		\includegraphics[width=\textwidth]{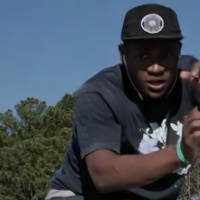}
		\caption{PaCNet (ours), PSNR = 36.11dB}
		\label{fig:skate-jump_f60_s20_covid_zoom}
		\vspace*{6pt}
	\end{subfigure}
	\caption{Denoising example with $\sigma = 20$. The figure shows frame 61 of the sequence \emph{skate-jump}. The PSNR values appearing in~\ref{fig:skate-jump_f60_s20_vnlb_rect},~\ref{fig:skate-jump_f60_s20_vnlnet_rect},~\ref{fig:skate-jump_f60_s20_fastdvd_rect}~and~\ref{fig:skate-jump_f60_s20_covid_rect} refer to the whole frame, whereas those in~\ref{fig:skate-jump_f60_s20_vnlb_zoom},~\ref{fig:skate-jump_f60_s20_vnlnet_zoom},~\ref{fig:skate-jump_f60_s20_fastdvd_zoom}~and~\ref{fig:skate-jump_f60_s20_covid_zoom} refer to the cropped area. As can be seen, PaCNet leads to better reconstructed result -- see the eyes and the details in the background trees.}
	\label{fig:skate-jump_f60_s20}
\end{figure*}

\begin{figure*}
    \centering
	\begin{subfigure}{0.312\textwidth}
	    \captionsetup{justification=centering}
		\includegraphics[width=\textwidth]{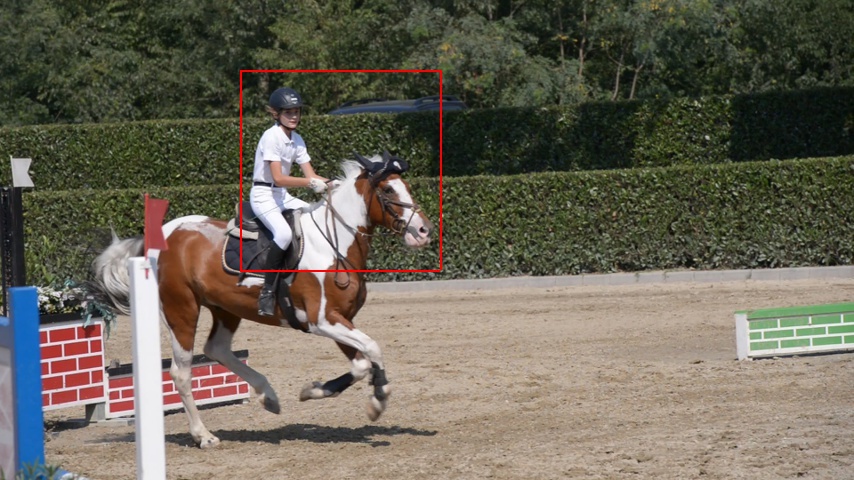}
		\caption{Original}
		\label{fig:horsejump-stick_f47_s40_c_rect}
		\vspace*{6pt}
	\end{subfigure}
	\begin{subfigure}{0.312\textwidth}
	    \captionsetup{justification=centering}
		\includegraphics[width=\textwidth]{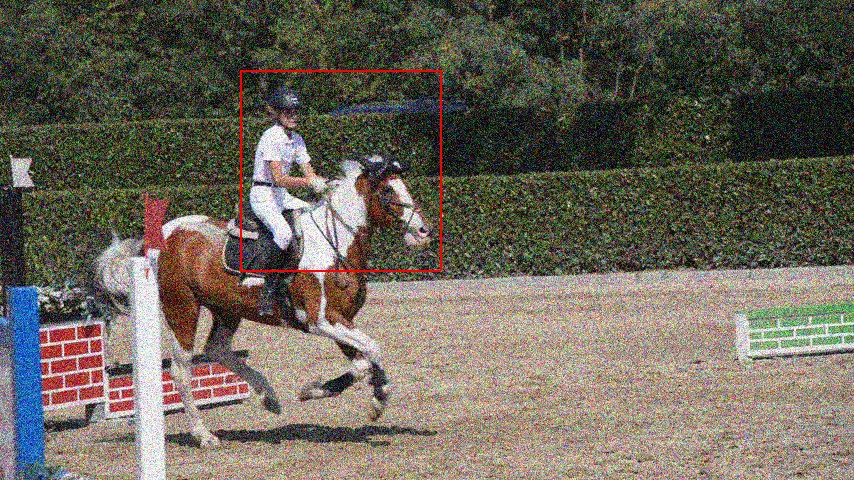}
		\caption{Noisy with $\sigma = 40$}
		\label{fig:horsejump-stick_f47_s40_n_rect}
		\vspace*{6pt}
	\end{subfigure}
	\begin{subfigure}{0.312\textwidth}
	    \captionsetup{justification=centering}
		\includegraphics[width=\textwidth]{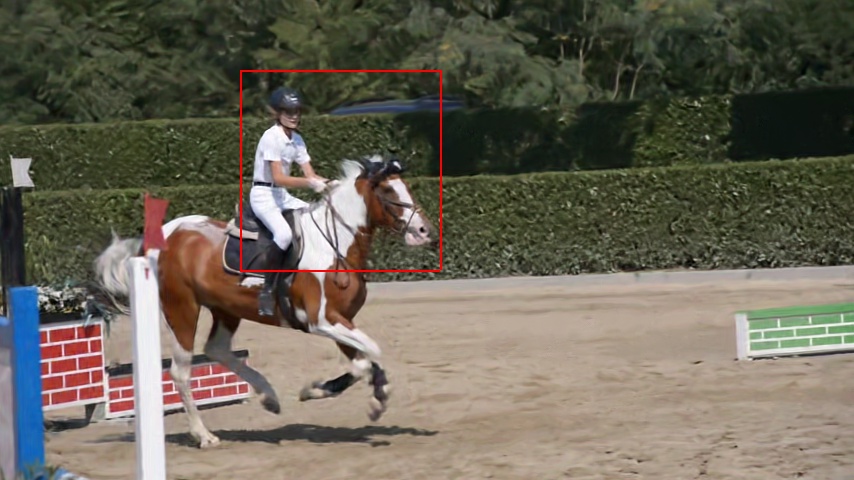}
		\caption{VNLB~\cite{arias2018video}, PSNR = 28.66dB}
		\label{fig:horsejump-stick_f47_s40_vnlb_rect}
		\vspace*{6pt}
	\end{subfigure}
	\begin{subfigure}{0.312\textwidth}
	    \captionsetup{justification=centering}
		\includegraphics[width=\textwidth]{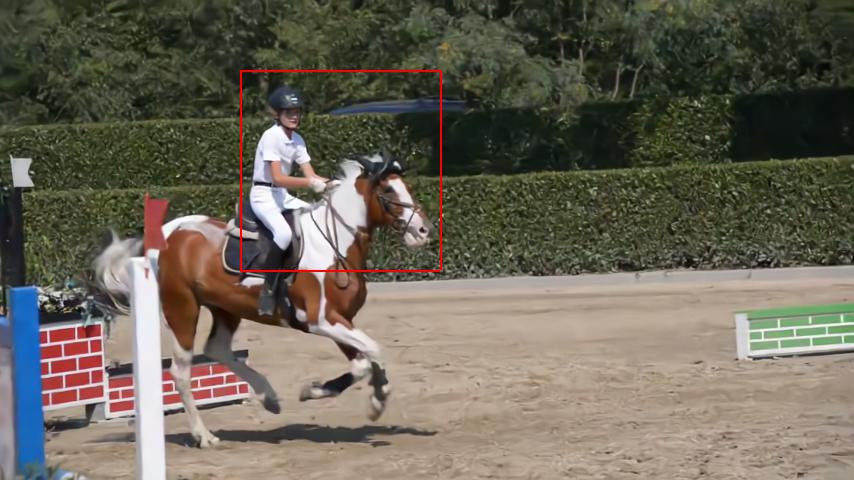}
		\caption{VNLnet~\cite{davy2019non}, PSNR = 29.03dB}
		\label{fig:horsejump-stick_f47_s40_vnlnet_rect}
		\vspace*{6pt}
	\end{subfigure}
	\begin{subfigure}{0.312\textwidth}
	    \captionsetup{justification=centering}
		\includegraphics[width=\textwidth]{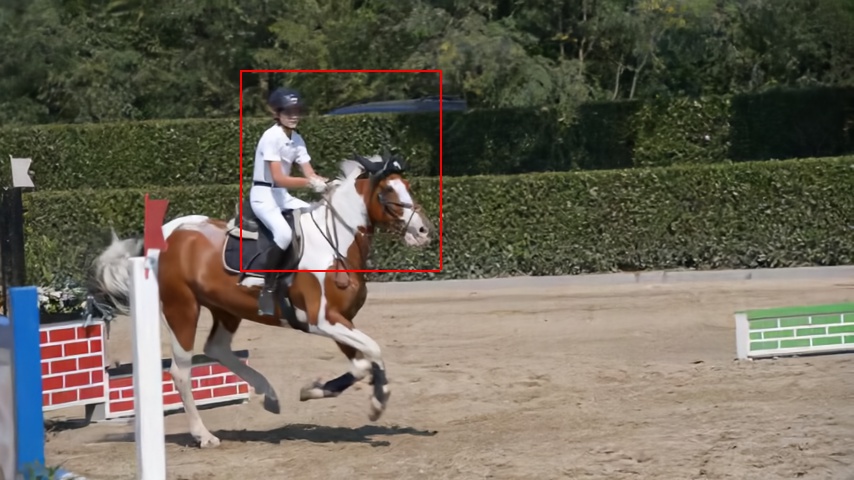}
		\caption{FastDVDnet~\cite{tassano2020fastdvdnet}, PSNR = 29.27dB}
		\label{fig:horsejump-stick_f47_s40_fastdvd_rect}
		\vspace*{6pt}
	\end{subfigure}
	\begin{subfigure}{0.312\textwidth}
	    \captionsetup{justification=centering}
		\includegraphics[width=\textwidth]{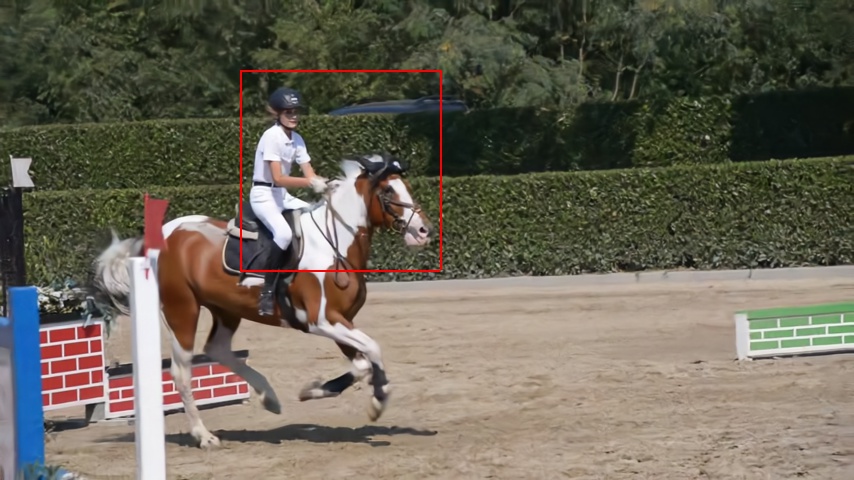}
		\caption{PaCNet (ours), PSNR = 29.73dB}
		\label{fig:horsejump-stick_f47_s40_covid_rect}
		\vspace*{6pt}
	\end{subfigure}
	\begin{subfigure}{0.312\textwidth}
	    \captionsetup{justification=centering}
		\includegraphics[width=\textwidth]{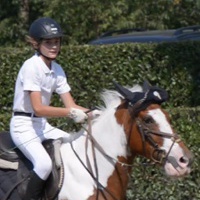}
		\caption{Original}
		\label{fig:horsejump-stick_f47_s40_c_zoom}
		\vspace*{6pt}
	\end{subfigure}
	\begin{subfigure}{0.312\textwidth}
	    \captionsetup{justification=centering}
		\includegraphics[width=\textwidth]{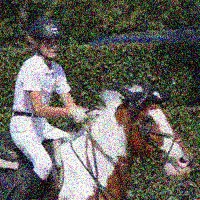}
		\caption{Noisy with $\sigma = 40$}
		\label{fig:horsejump-stick_f47_s40_n_zoom}
		\vspace*{6pt}
	\end{subfigure}
	\begin{subfigure}{0.312\textwidth}
	    \captionsetup{justification=centering}
		\includegraphics[width=\textwidth]{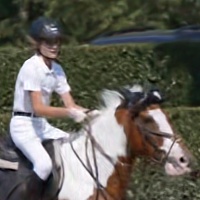}
		\caption{VNLB~\cite{arias2018video}, PSNR = 27.92dB}
		\label{fig:horsejump-stick_f47_s40_vnlb_zoom}
		\vspace*{6pt}
	\end{subfigure}
	\begin{subfigure}{0.312\textwidth}
	    \captionsetup{justification=centering}
		\includegraphics[width=\textwidth]{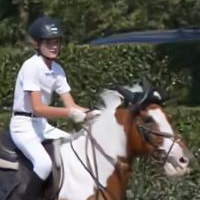}
		\caption{VNLnet~\cite{davy2019non}, PSNR = 27.95dB}
		\label{fig:horsejump-stick_f47_s40_vnlnet_zoom}
		\vspace*{6pt}
	\end{subfigure}
	\begin{subfigure}{0.312\textwidth}
	    \captionsetup{justification=centering}
		\includegraphics[width=\textwidth]{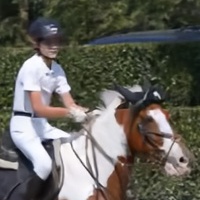}
		\caption{FastDVDnet~\cite{tassano2020fastdvdnet}, PSNR = 28.23dB}
		\label{fig:horsejump-stick_f47_s40_fastdvd_zoom}
		\vspace*{6pt}
	\end{subfigure}
	\begin{subfigure}{0.312\textwidth}
	    \captionsetup{justification=centering}
		\includegraphics[width=\textwidth]{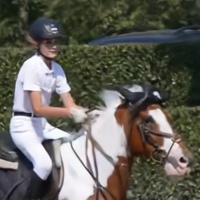}
		\caption{PaCNet (ours), PSNR = 29.07dB}
		\label{fig:horsejump-stick_f47_s40_covid_zoom}
		\vspace*{6pt}
	\end{subfigure}
	\caption{Denoising example with $\sigma = 40$. The figure shows frame 48 of the sequence \emph{horsejump-stick}. The PSNR values appearing in~\ref{fig:horsejump-stick_f47_s40_vnlb_rect},~\ref{fig:horsejump-stick_f47_s40_vnlnet_rect},~\ref{fig:horsejump-stick_f47_s40_fastdvd_rect}~and~\ref{fig:horsejump-stick_f47_s40_covid_rect} refer to the whole frame, whereas those in~\ref{fig:horsejump-stick_f47_s40_vnlb_zoom},~\ref{fig:horsejump-stick_f47_s40_vnlnet_zoom},~\ref{fig:horsejump-stick_f47_s40_fastdvd_zoom}~and~\ref{fig:horsejump-stick_f47_s40_covid_zoom} refer to the cropped area. As can be seen, PaCNet leads to better reconstructed results -- see the face and the details in the background shrubs.}
	\label{fig:horsejump-stick_f47_s40}
\end{figure*}

Figures~\ref{fig:skate-jump_f60_s20}~and~\ref{fig:horsejump-stick_f47_s40} present qualitative comparisons of our algorithm with leading alternatives. As can be seen, our method reconstructs video frames more faithfully than the competing algorithms. For example, in Figure~\ref{fig:skate-jump_f60_s20}, PaCNet manages to recover the eyes and preserves more details in the background trees. The comparison with VNLB~\cite{arias2018video} and FastDVDnet~\cite{tassano2020fastdvdnet} shows that PaCNet tends to produce sharper frames with more details. The strength of the FastDVDnet is its reliance on a plain CNN architecture, but its weakness is lack of explicitly harnessing non-local self-similarity. In contrast, while VNLB leverages non-local redundancy, it is still inferior to a supervised trained CNN. PaCNet enjoys both worlds, as it combines a CNN processing with leveraging of non-local self-similarity. The synergy between these two leads to SOTA denoising performance both visually and in terms of PSNR. 

We also compare PaCNet with VNLnet~\cite{davy2019non}, which combines nearest neighbor search with CNN for video denoising. Although VNLnet performs non-local filtering, its effectiveness is limited due to the restricted use of central pixels of patches. As can be seen in Figure~\ref{fig:skate-jump_f60_s20_vnlnet_zoom},  VNLnet creates a sharp frame with a good recovery, but suffers from artifacts along  edges, as reflected by a 2.5dB drop compared to PaCNet in Figure~\ref{fig:skate-jump_f60_s20_covid_zoom}. In Figure~\ref{fig:skate-jump_f60_s20_vnlnet_zoom}, these artifacts can be seen on the man's cap. We bring more qualitative comparisons in the supplementary material. Beyond high PSNR and the sharp reconstructed frames, our method produces video sequences with low flickering -- this can be seen in video sequences in the supplementary material.
\footnotetext[2]{ FastDVDnet~\cite{tassano2020fastdvdnet} PSNR values are obtained from the released code. The rest of the values reported in Tables~\ref{tab:denoising_known_sig}~and~\ref{tab:denoising_clipped} are taken from~\cite{tassano2020fastdvdnet}. }
\footnotetext[3]{ The PSNR value for VNLnet~\cite{davy2019non} with $\sigma = 30$ is missing as~\cite{davy2019non} did not provide a model for this noise level. }

In addition to the above, we evaluate the denoising performance of PaCNet in experiments with a clipped Gaussian noise (i.e., truncation of noisy pixels to $[0, 1]$) and provide a comprehensive comparison to recent SOTA algorithms for this type of distortion, ViDeNN~\cite{claus2019videnn} and FastDVDnet~\cite{tassano2020fastdvdnet}. In this experiment we use the same network parameters and the same training and test sets as above. Average PSNR results are reported in Table~\ref{tab:denoising_clipped}, exposing a similar trend to previous experiments. More specifically, PaCNet shows considerable improvement in PSNR of {0.8~dB - 1.4~dB}, while the improvement decreases with an increase of noise level.

\subsection{Single Image Denoising}
The proposed algorithm can be easily reduced to a single image denoiser by omitting the temporal denoising network and setting $T_s=0$. We refer to this as S-CNN-0. Among existing image denoisers, this configuration is most similar to LIDIA~\cite{vaksman2020lidia}, as both methods perform nearest neighbor search as an augmentation for denoising. Table~\ref{tab:image_denoising} shows that these two denoisers have very similar performance. In order to demonstrate the impact of each component of PaCNet, we add two columns to Table~\ref{tab:denoising_known_sig}: S-CNN-0 and S-CNN-3. S-CNN-0 reports the performance of our scheme in an intra-frame denoising configuration, in which the video is denoised frame by frame independently. S-CNN-3 shows the PaCNet performance without the temporal filtering. In this scenario, we set $T_s = 3$, extending the nearest neighbor search in $7$ adjacent frames. As can be seen, extending the nearest neighbor search to nearby frames gains more than 1~dB in PSNR, compared to a frame by frame denoising. Temporal filtering adds 0.15-0.8~dB, where this benefit increases with the increase in noise level. In addition, the temporal filter plays a key role in the reduction of flickering.  


\section{Conclusion}
This work presents a novel algorithm for video denoising. Our method augments the processed video with patch-craft frames and applies spatial and temporal filtering on the enlarged sequence. The augmentation leverages non-local redundancy, similar to the way the patch-based framework operates. The spatial denoising network consists of separable convolutional layers, which allow for reasonable memory and computational complexities. The temporal CNN reduces flickering by imposing temporal continuity. We demonstrate the proposed method in extensive tests.\footnote{The code reproducing the results of this paper is available at \\ \href{https://github.com/grishavak/PaCNet-denoiser}{https://github.com/grishavak/PaCNet-denoiser}.}


{\small
\bibliographystyle{ieee_fullname}
\bibliography{egbib}
}

\clearpage
\begin{appendices}
\section{Difference Between Patch-Craft Frames}
\begin{figure}[h!]
    \centering
	\begin{subfigure}{0.46\textwidth}
	    \captionsetup{justification=centering}
		\includegraphics[width=\textwidth]{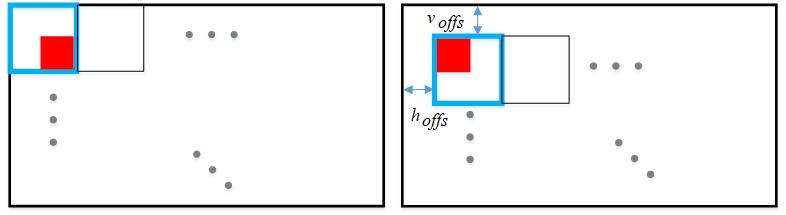}
	\end{subfigure}
	\caption{Inconsistency of patch overlaps in patch-craft frames with different offsets.} 
	\label{fig:patch-craft_inconsistency}
\end{figure}

As described in Section~\ref{sec:patch_craft_frames}, our algorithm augments each processed frame with $nf$ patch-craft frames. We emphasize that \emph{all these are different}, not identical nor shifted versions of each other. Thus, each brings an important additional information for the denoising to leverage. 
Here is an illustrative example to clarify this point. Assume for simplicity that $n = 1$, i.e., only one nearest neighbor is used. Consider two patch-craft frames with two different offsets, $[0,0]$ and $[h_{offs}, v_{offs}]$, as shown in Figure~\ref{fig:patch-craft_inconsistency}.

\noindent Consider the blue patches in these frames and their overlap red area. Each blue patch is a nearest neighbor (NN) of a corresponding patch in the processed frame, which means that the blue patches come from different locations in the video (perhaps even different frames). As such, their red regions are different, holding each additional information about the corresponding area of the processed frame. 
More broadly, all patch-craft frames are similar to each other but not identical, thus enriching the denoising process.

\section{Additional Details Regarding Training}
\begin{figure}
    \centering
	\begin{subfigure}{0.4\textwidth}
	    \captionsetup{justification=centering}
		\includegraphics[width=\textwidth]{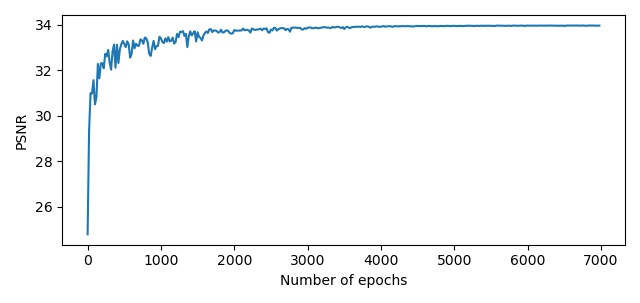}
		\caption{S-CNN}
		\label{fig:training:s_cnn}
	\end{subfigure}
	\begin{subfigure}{0.4\textwidth}
	    \captionsetup{justification=centering}
		\includegraphics[width=\textwidth]{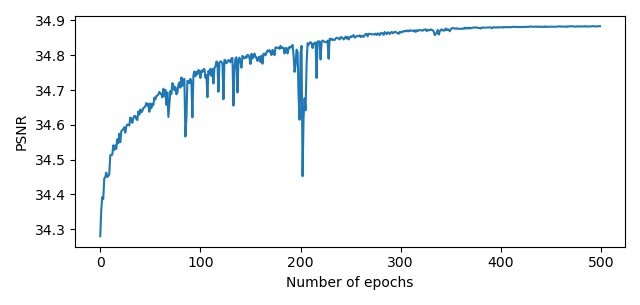}
		\caption{T-CNN}
		\label{fig:training:t_cnn}
	\end{subfigure}
	\caption{PSNR vs. the number of epochs for the validation set during training of the spatial and the temporal denoising networks, S-CNN and T-CNN, for noise level $\sigma = 30$. (We use different validation sets for each network).} 
	\label{fig:training}
\end{figure}

Figures~\ref{fig:training:s_cnn}~and~\ref{fig:training:t_cnn} present graphs of PSNR versus number of epochs during training of our networks. The values shown in the graphs are a rough estimation of training PSNR obtained by evaluating the networks on a small set of short videos randomly cropped from the training set. The spatial network, S-CNN, is trained using spatio-temporal $3D$ boxes of size $150 \times 150 \times 7$, applying denoising on the central frame of size $64 \times 64 \times 1$, where the rest of the box is used for nearest neighbor search. The boxes are randomly cropped from the training video seqiences. We use batches of size 10 and train the network for $7000$ epochs. For training the temporal network, T-CNN, we use batches of 10 randomly cropped spatial-temporal boxes of size $64 \times 64 \times 7$ and run training for $500$ epochs.

\section{Additional Results}
\begin{figure}
    \centering
    \captionsetup{justification=centering}
	\includegraphics[width=0.33\textwidth]{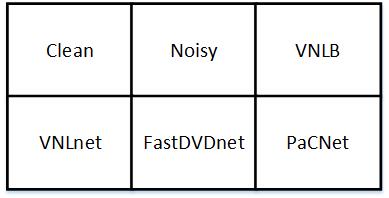}
	\caption{Video chart.}
	\label{fig:video_chart}
\end{figure}

\begin{figure*}
    \centering
	\begin{subfigure}{0.33\textwidth}
	    \captionsetup{justification=centering}
		\includegraphics[width=\textwidth]{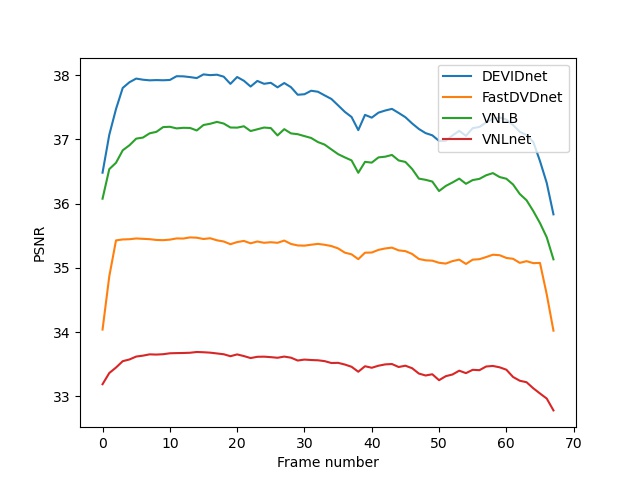}
		\caption{\emph{skate-jump}}
		\label{fig:fr_psnr:skate-jump_20}
	\end{subfigure}
	\begin{subfigure}{0.33\textwidth}
	    \captionsetup{justification=centering}
		\includegraphics[width=\textwidth]{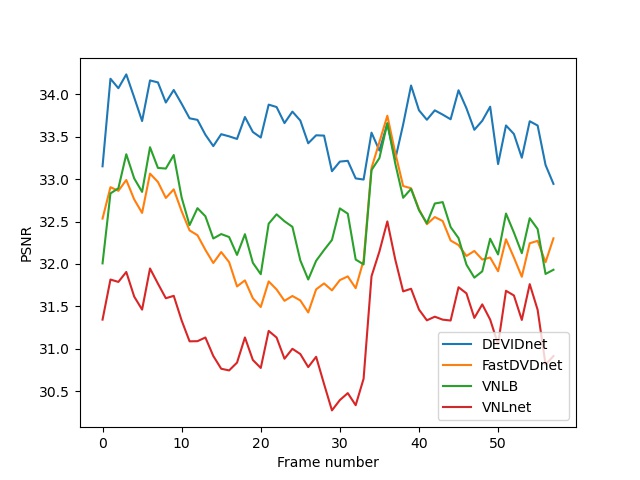}
		\caption{\emph{horsejump-stick}}
		\label{fig:fr_psnr:horsejump-stick_20}
	\end{subfigure}
	\begin{subfigure}{0.33\textwidth}
	    \captionsetup{justification=centering}
		\includegraphics[width=\textwidth]{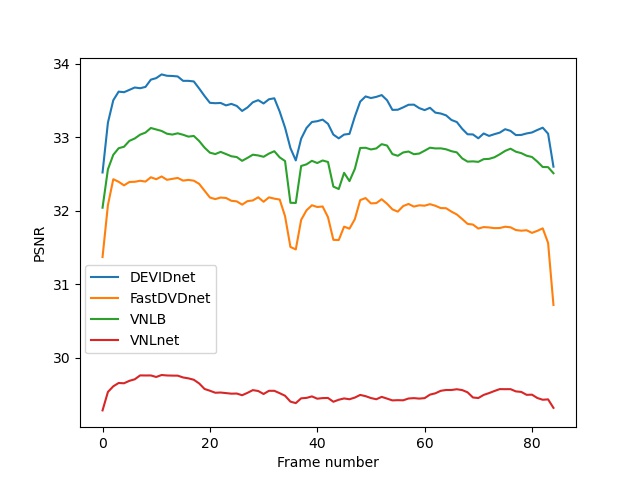}
		\caption{\emph{salsa}}
		\label{fig:fr_psnr:salsa_20}
	\end{subfigure}
	\begin{subfigure}{0.33\textwidth}
	    \captionsetup{justification=centering}
		\includegraphics[width=\textwidth]{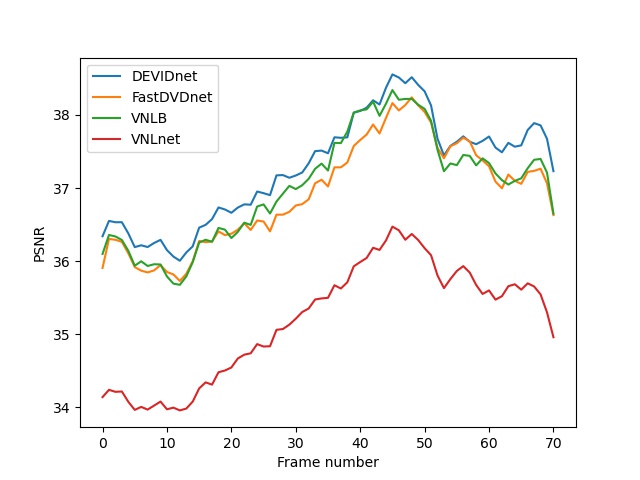}
		\caption{\emph{aerobatics}}
		\label{fig:fr_psnr:aerobatics_20}
	\end{subfigure}
	\begin{subfigure}{0.33\textwidth}
	    \captionsetup{justification=centering}
		\includegraphics[width=\textwidth]{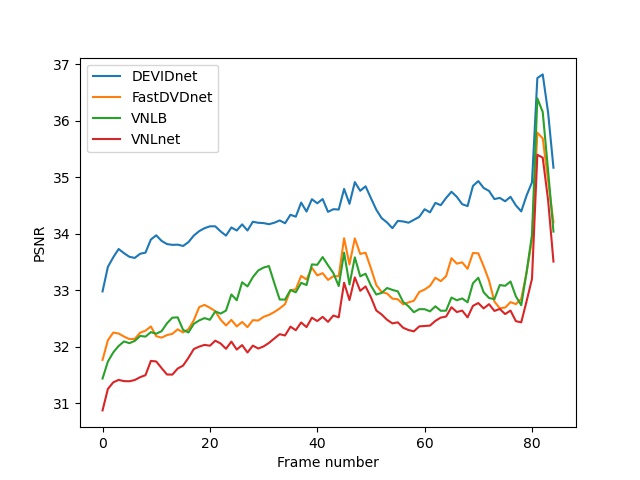}
		\caption{\emph{girl-dog}}
		\label{fig:fr_psnr:girl-dog_20}
	\end{subfigure}
	\begin{subfigure}{0.33\textwidth}
	    \captionsetup{justification=centering}
		\includegraphics[width=\textwidth]{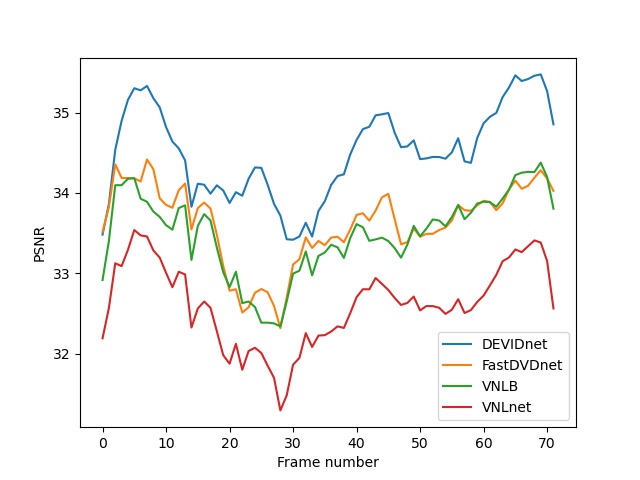}
		\caption{\emph{tandem}}
		\label{fig:fr_psnr:tandem_20}
	\end{subfigure}
	\begin{subfigure}{0.33\textwidth}
	    \captionsetup{justification=centering}
		\includegraphics[width=\textwidth]{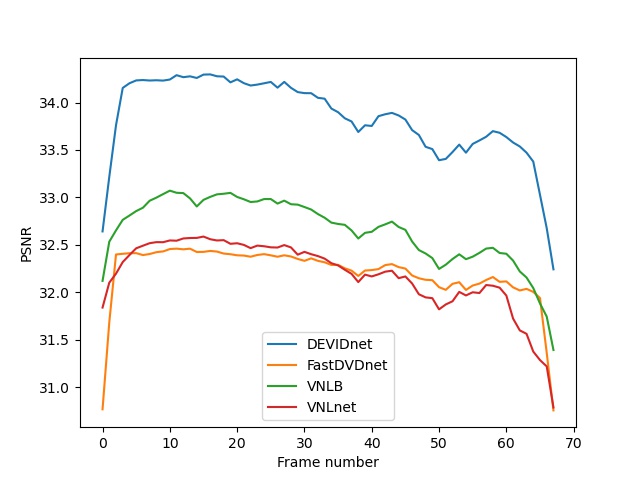}
		\caption{\emph{skate-jump}}
		\label{fig:fr_psnr:skate-jump_40}
	\end{subfigure}
	\begin{subfigure}{0.33\textwidth}
	    \captionsetup{justification=centering}
		\includegraphics[width=\textwidth]{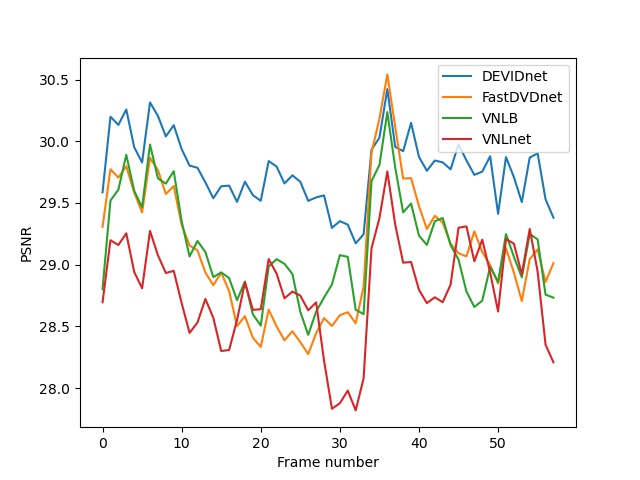}
		\caption{\emph{horsejump-stick}}
		\label{fig:fr_psnr:horsejump-stick_40}
	\end{subfigure}
	\begin{subfigure}{0.33\textwidth}
	    \captionsetup{justification=centering}
		\includegraphics[width=\textwidth]{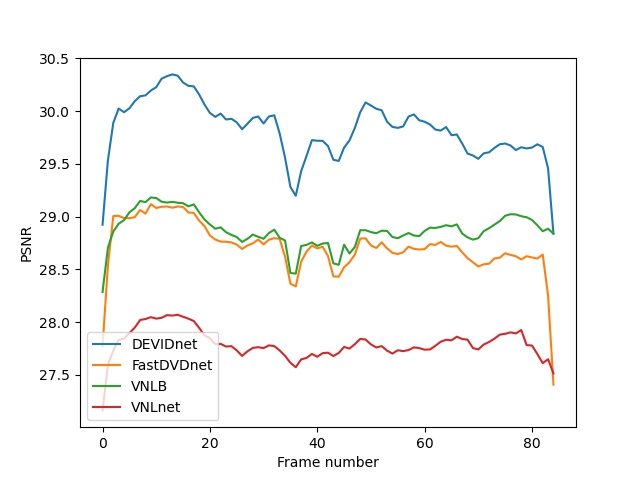}
		\caption{\emph{salsa}}
		\label{fig:fr_psnr:salsa_40}
	\end{subfigure}
	\begin{subfigure}{0.33\textwidth}
	    \captionsetup{justification=centering}
		\includegraphics[width=\textwidth]{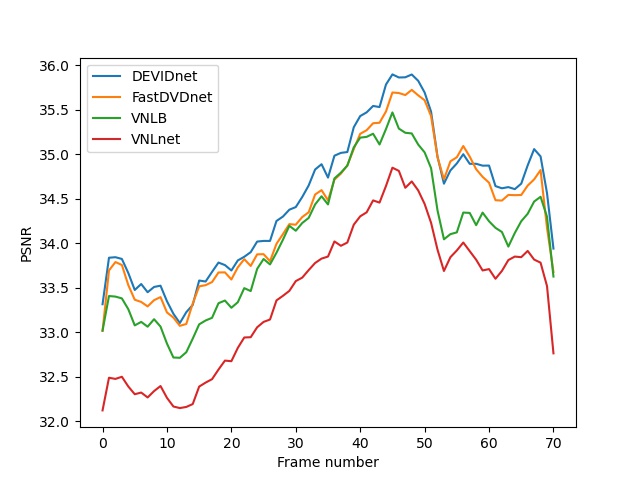}
		\caption{\emph{aerobatics}}
		\label{fig:fr_psnr:aerobatics_40}
	\end{subfigure}
	\begin{subfigure}{0.33\textwidth}
	    \captionsetup{justification=centering}
		\includegraphics[width=\textwidth]{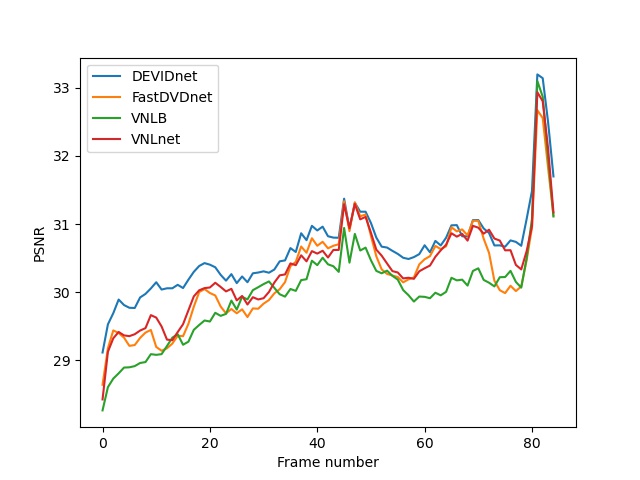}
		\caption{\emph{girl-dog}}
		\label{fig:fr_psnr:girl-dog_40}
	\end{subfigure}
	\begin{subfigure}{0.33\textwidth}
	    \captionsetup{justification=centering}
		\includegraphics[width=\textwidth]{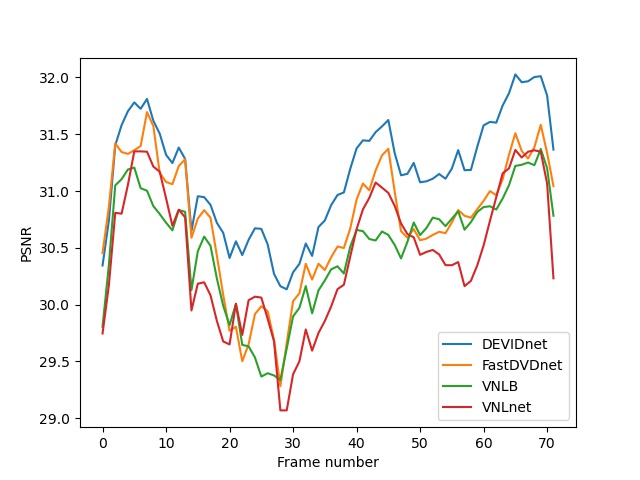}
		\caption{\emph{tandem}}
		\label{fig:fr_psnr:tandem_40}
	\end{subfigure}
	\caption{PSNR vs. frame number for video sequences \emph{skate-jump}, \emph{horsejump-stick}, \emph{salsa}, \emph{aerobatics}, \emph{girl-dog}, and \emph{tandem}. The two first rows show denoising experiments with noise level $\sigma = 20$ and the third and fourth rows with $\sigma = 40$.} 
	\label{fig:fr_psnr}
\end{figure*}

\begin{figure*}
    \centering
	\begin{subfigure}{0.32\textwidth}
	    \captionsetup{justification=centering}
		\includegraphics[width=\textwidth]{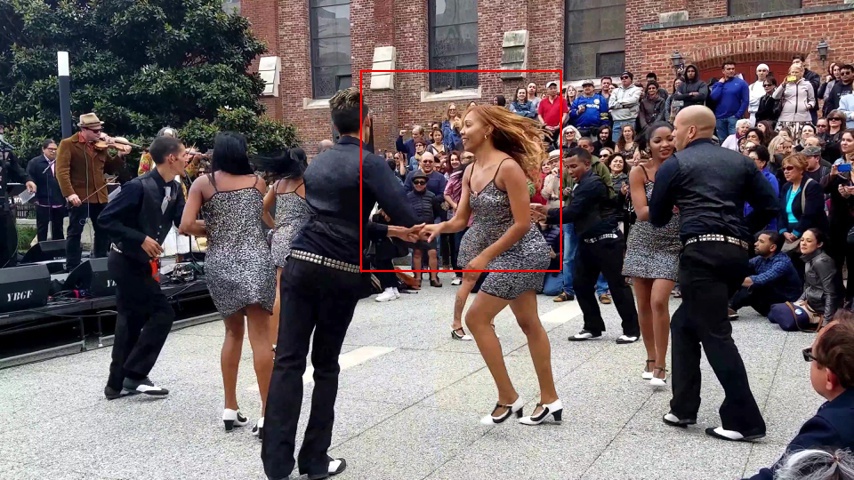}
		\caption{Original}
		\label{fig:salsa_f8_s40_c_rect}
		\vspace*{6pt}
	\end{subfigure}
	\begin{subfigure}{0.32\textwidth}
	    \captionsetup{justification=centering}
		\includegraphics[width=\textwidth]{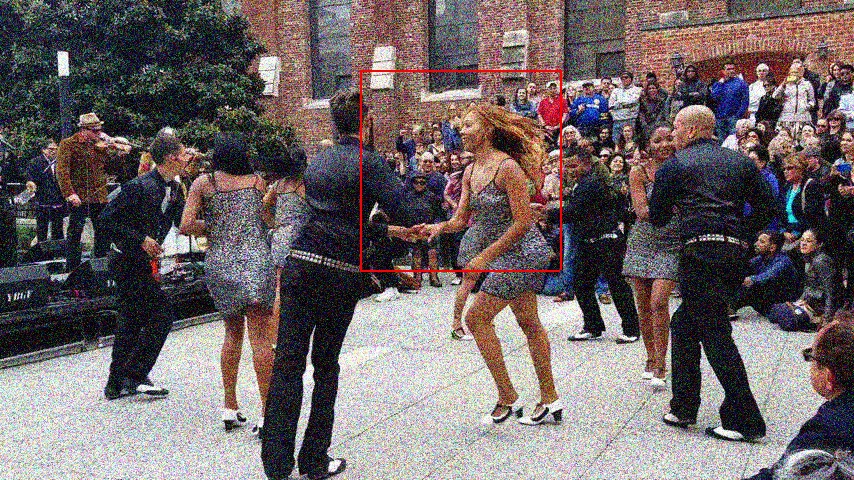}
		\caption{Noisy with $\sigma = 40$}
		\label{fig:salsa_f8_s40_n_rect}
		\vspace*{6pt}
	\end{subfigure}
	\begin{subfigure}{0.32\textwidth}
	    \captionsetup{justification=centering}
		\includegraphics[width=\textwidth]{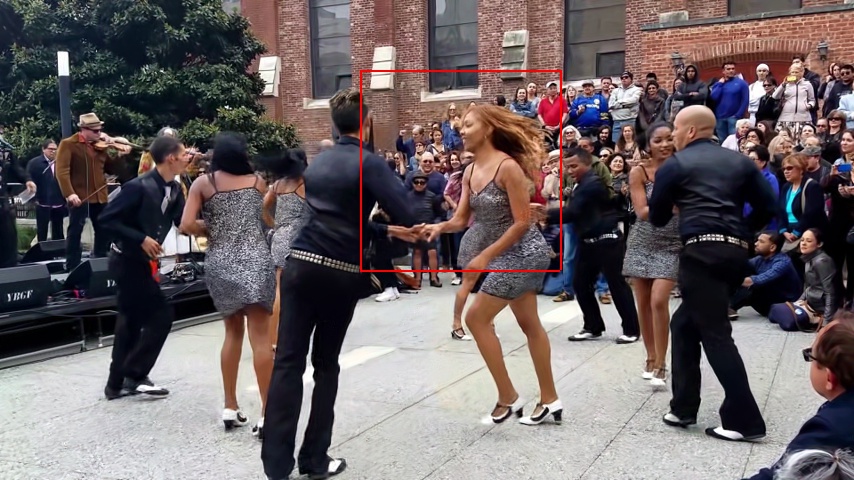}
		\caption{VNLB~\cite{arias2018video}, PSNR = 29.14dB}
		\label{fig:salsa_f8_s40_vnlb_rect}
		\vspace*{6pt}
	\end{subfigure}
	\begin{subfigure}{0.32\textwidth}
	    \captionsetup{justification=centering}
		\includegraphics[width=\textwidth]{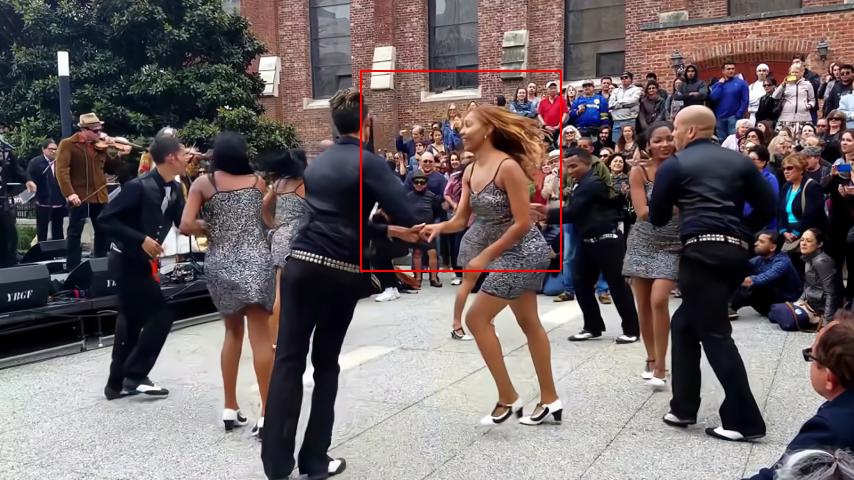}
		\caption{VNLnet~\cite{davy2019non}, PSNR = 28.03dB}
		\label{fig:salsa_f8_s40_vnlnet_rect}
		\vspace*{6pt}
	\end{subfigure}
	\begin{subfigure}{0.32\textwidth}
	    \captionsetup{justification=centering}
		\includegraphics[width=\textwidth]{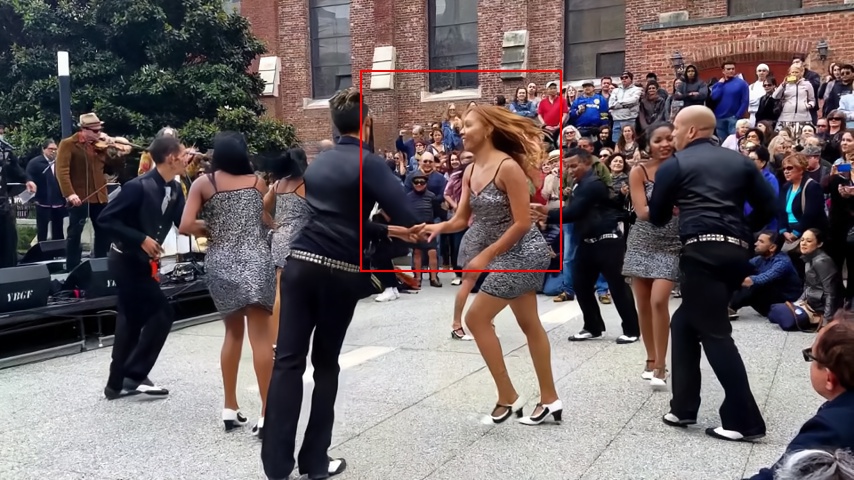}
		\caption{FastDVDnet~\cite{tassano2020fastdvdnet}, PSNR = 29.03dB}
		\label{fig:salsa_f8_s40_fastdvd_rect}
		\vspace*{6pt}
	\end{subfigure}
	\begin{subfigure}{0.32\textwidth}
	    \captionsetup{justification=centering}
		\includegraphics[width=\textwidth]{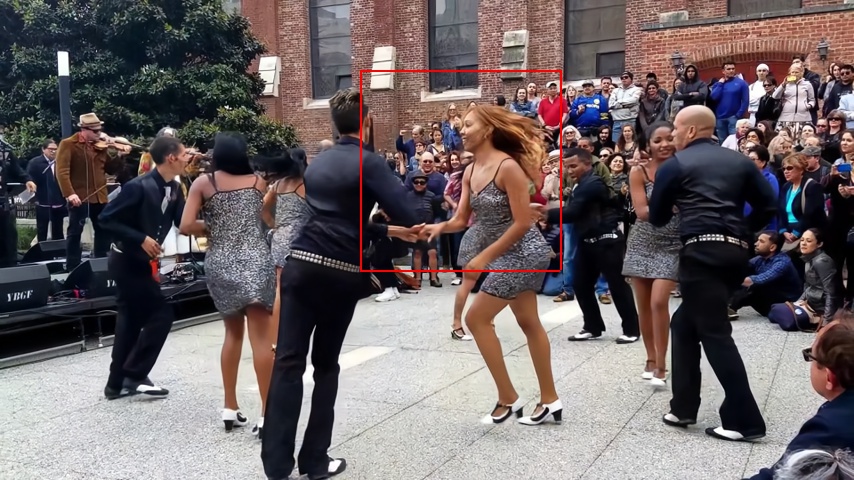}
		\caption{PaCNet (ours), PSNR = 30.15dB}
		\label{fig:salsa_f8_s40_covid_rect}
		\vspace*{6pt}
	\end{subfigure}
	\begin{subfigure}{0.32\textwidth}
	    \captionsetup{justification=centering}
		\includegraphics[width=\textwidth]{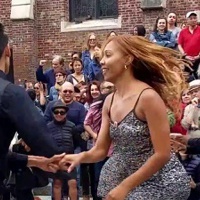}
		\caption{Original}
		\label{fig:salsa_f8_s40_c_zoom}
		\vspace*{6pt}
	\end{subfigure}
	\begin{subfigure}{0.32\textwidth}
	    \captionsetup{justification=centering}
		\includegraphics[width=\textwidth]{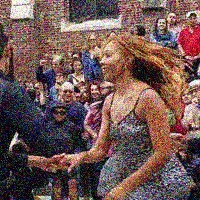}
		\caption{Noisy with $\sigma = 40$}
		\label{fig:salsa_f8_s40_n_zoom}
		\vspace*{6pt}
	\end{subfigure}
	\begin{subfigure}{0.32\textwidth}
	    \captionsetup{justification=centering}
		\includegraphics[width=\textwidth]{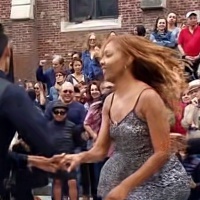}
		\caption{VNLB~\cite{arias2018video}, PSNR = 27.68dB}
		\label{fig:salsa_f8_s40_vnlb_zoom}
		\vspace*{6pt}
	\end{subfigure}
	\begin{subfigure}{0.32\textwidth}
	    \captionsetup{justification=centering}
		\includegraphics[width=\textwidth]{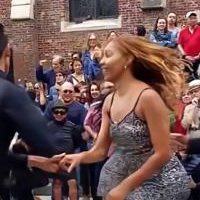}
		\caption{VNLnet~\cite{davy2019non}, PSNR = 26.69dB}
		\label{fig:salsa_f8_s40_vnlnet_zoom}
		\vspace*{6pt}
	\end{subfigure}
	\begin{subfigure}{0.32\textwidth}
	    \captionsetup{justification=centering}
		\includegraphics[width=\textwidth]{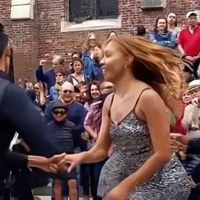}
		\caption{FastDVDnet~\cite{tassano2020fastdvdnet}, PSNR = 27.56dB}
		\label{fig:salsa_f8_s40_fastdvd_zoom}
		\vspace*{6pt}
	\end{subfigure}
	\begin{subfigure}{0.32\textwidth}
	    \captionsetup{justification=centering}
		\includegraphics[width=\textwidth]{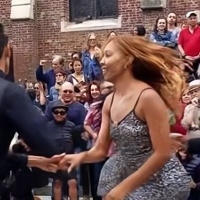}
		\caption{PaCNet (ours), PSNR = 28.31dB}
		\label{fig:salsa_f8_s40_covid_zoom}
		\vspace*{6pt}
	\end{subfigure}
	\caption{Denoising example with $\sigma = 40$. The figure shows frame 9 of the sequence \emph{salsa}. The PSNR values appearing in~\ref{fig:salsa_f8_s40_vnlb_rect},~\ref{fig:salsa_f8_s40_vnlnet_rect},~\ref{fig:salsa_f8_s40_fastdvd_rect}~and~\ref{fig:salsa_f8_s40_covid_rect} refer to the whole frame, whereas those in~\ref{fig:salsa_f8_s40_vnlb_zoom},~\ref{fig:salsa_f8_s40_vnlnet_zoom},~\ref{fig:salsa_f8_s40_fastdvd_zoom}~and~\ref{fig:salsa_f8_s40_covid_zoom} refer to the cropped area. As can be seen, PaCNet leads to better reconstructed results -- see the face and the details in the background building.}
	\label{fig:salsa_f8_s40}
\end{figure*}

\begin{figure*}
    \centering
	\begin{subfigure}{0.32\textwidth}
	    \captionsetup{justification=centering}
		\includegraphics[width=\textwidth]{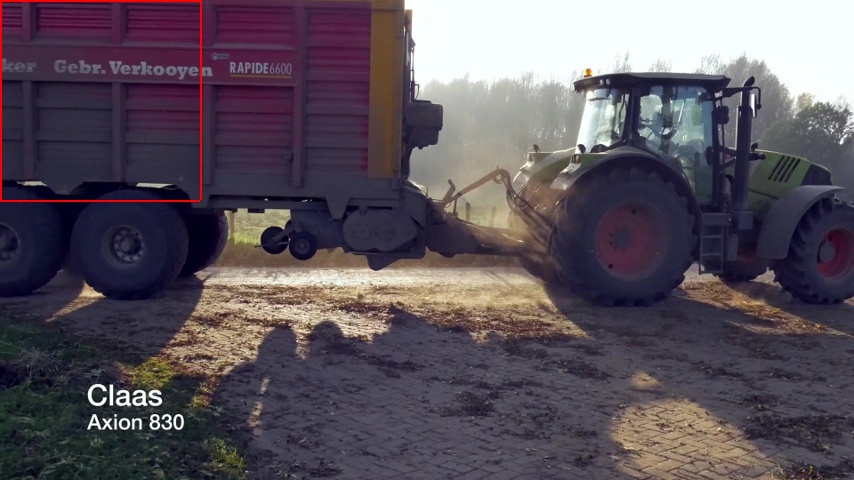}
		\caption{Original}
		\label{fig:tractor_f22_s20_c_rect}
		\vspace*{6pt}
	\end{subfigure}
	\begin{subfigure}{0.32\textwidth}
	    \captionsetup{justification=centering}
		\includegraphics[width=\textwidth]{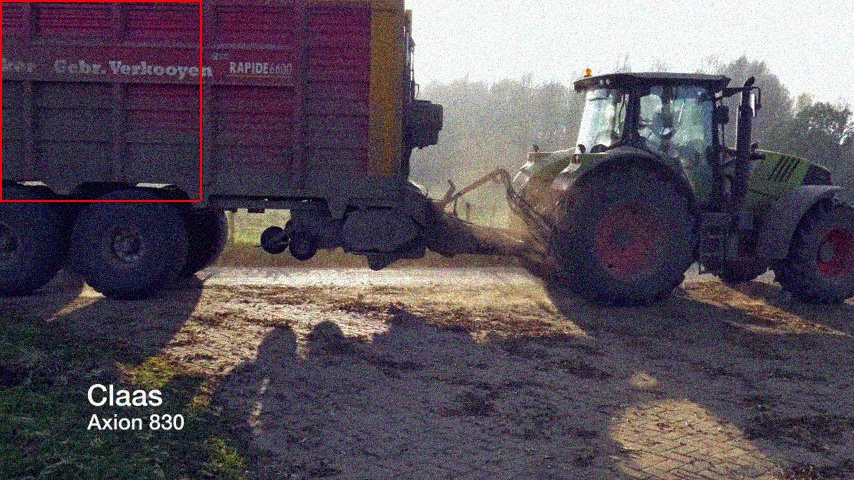}
		\caption{Noisy with $\sigma = 20$}
		\label{fig:tractor_f22_s20_n_rect}
		\vspace*{6pt}
	\end{subfigure}
	\begin{subfigure}{0.32\textwidth}
	    \captionsetup{justification=centering}
		\includegraphics[width=\textwidth]{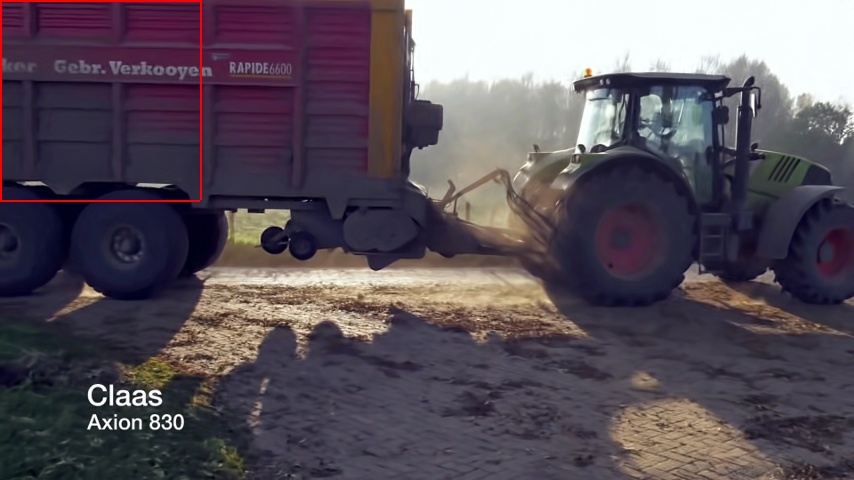}
		\caption{VNLB~\cite{arias2018video}, PSNR = 34.18dB}
		\label{fig:tractor_f22_s20_vnlb_rect}
		\vspace*{6pt}
	\end{subfigure}
	\begin{subfigure}{0.32\textwidth}
	    \captionsetup{justification=centering}
		\includegraphics[width=\textwidth]{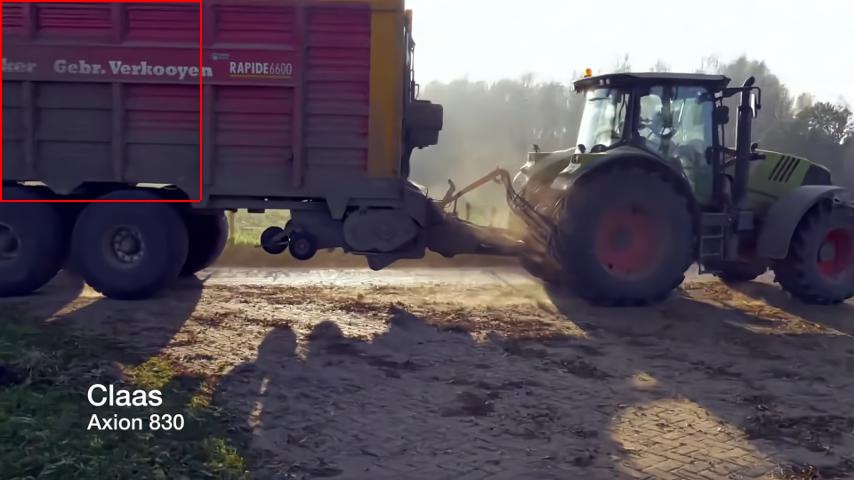}
		\caption{VNLnet~\cite{davy2019non}, PSNR = 33.26dB}
		\label{fig:tractor_f22_s20_vnlnet_rect}
		\vspace*{6pt}
	\end{subfigure}
	\begin{subfigure}{0.32\textwidth}
	    \captionsetup{justification=centering}
		\includegraphics[width=\textwidth]{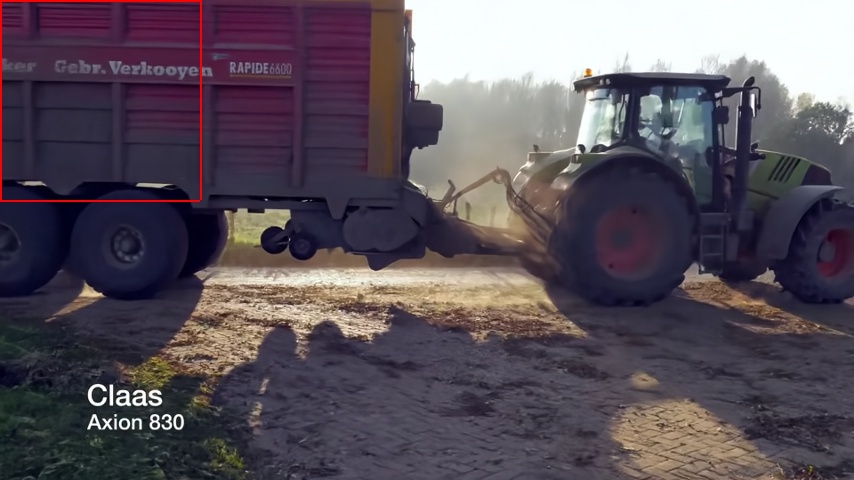}
		\caption{FastDVDnet~\cite{tassano2020fastdvdnet}, PSNR = 34.42dB}
		\label{fig:tractor_f22_s20_fastdvd_rect}
		\vspace*{6pt}
	\end{subfigure}
	\begin{subfigure}{0.32\textwidth}
	    \captionsetup{justification=centering}
		\includegraphics[width=\textwidth]{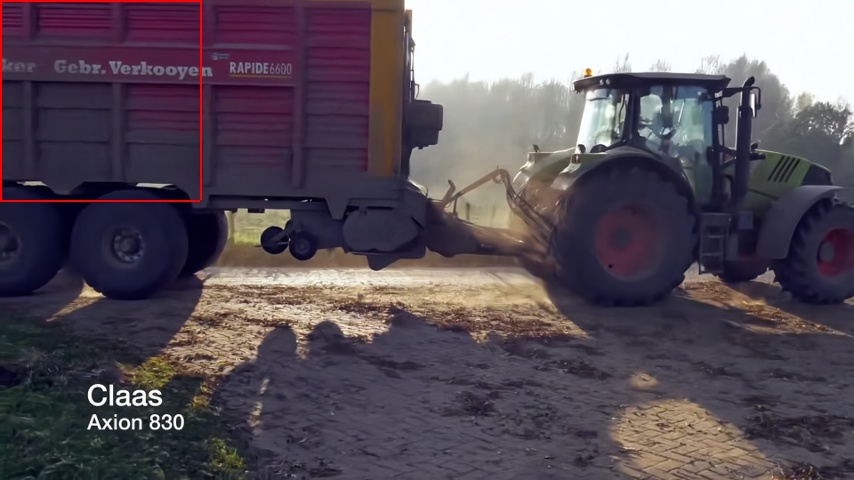}
		\caption{PaCNet (ours), PSNR = 34.74dB}
		\label{fig:tractor_f22_s20_covid_rect}
		\vspace*{6pt}
	\end{subfigure}
	\begin{subfigure}{0.32\textwidth}
	    \captionsetup{justification=centering}
		\includegraphics[width=\textwidth]{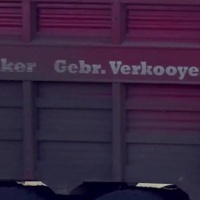}
		\caption{Original}
		\label{fig:tractor_f22_s20_c_zoom}
		\vspace*{6pt}
	\end{subfigure}
	\begin{subfigure}{0.32\textwidth}
	    \captionsetup{justification=centering}
		\includegraphics[width=\textwidth]{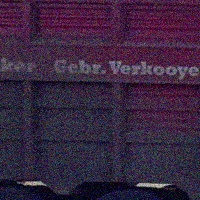}
		\caption{Noisy with $\sigma = 20$}
		\label{fig:tractor_f22_s20_n_zoom}
		\vspace*{6pt}
	\end{subfigure}
	\begin{subfigure}{0.32\textwidth}
	    \captionsetup{justification=centering}
		\includegraphics[width=\textwidth]{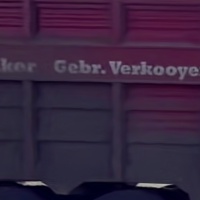}
		\caption{VNLB~\cite{arias2018video}, PSNR = 38.43dB}
		\label{fig:tractor_f22_s20_vnlb_zoom}
		\vspace*{6pt}
	\end{subfigure}
	\begin{subfigure}{0.32\textwidth}
	    \captionsetup{justification=centering}
		\includegraphics[width=\textwidth]{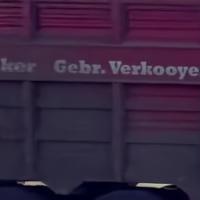}
		\caption{VNLnet~\cite{davy2019non}, PSNR = 37.71dB}
		\label{fig:tractor_f22_s20_vnlnet_zoom}
		\vspace*{6pt}
	\end{subfigure}
	\begin{subfigure}{0.32\textwidth}
	    \captionsetup{justification=centering}
		\includegraphics[width=\textwidth]{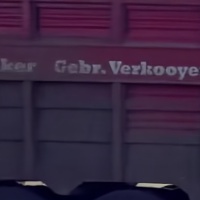}
		\caption{FastDVDnet~\cite{tassano2020fastdvdnet}, PSNR = 38.53dB}
		\label{fig:tractor_f22_s20_fastdvd_zoom}
		\vspace*{6pt}
	\end{subfigure}
	\begin{subfigure}{0.32\textwidth}
	    \captionsetup{justification=centering}
		\includegraphics[width=\textwidth]{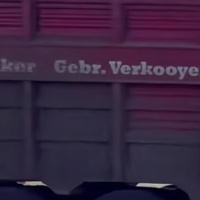}
		\caption{PaCNet (ours), PSNR = 39.09dB}
		\label{fig:tractor_f22_s20_covid_zoom}
		\vspace*{6pt}
	\end{subfigure}
	\caption{Denoising example with $\sigma = 20$. The figure shows frame 23 of the sequence \emph{tractor}. The PSNR values appearing in~\ref{fig:tractor_f22_s20_vnlb_rect},~\ref{fig:tractor_f22_s20_vnlnet_rect},~\ref{fig:tractor_f22_s20_fastdvd_rect}~and~\ref{fig:tractor_f22_s20_covid_rect} refer to the whole frame, whereas those in~\ref{fig:tractor_f22_s20_vnlb_zoom},~\ref{fig:tractor_f22_s20_vnlnet_zoom},~\ref{fig:tractor_f22_s20_fastdvd_zoom}~and~\ref{fig:tractor_f22_s20_covid_zoom} refer to the cropped area. As can be seen, PaCNet leads to better reconstructed results -- see the text on the trailer.}
	\label{fig:tractor_f22_s20}
\end{figure*}

\begin{figure*}
    \centering
	\begin{subfigure}{0.32\textwidth}
	    \captionsetup{justification=centering}
		\includegraphics[width=\textwidth]{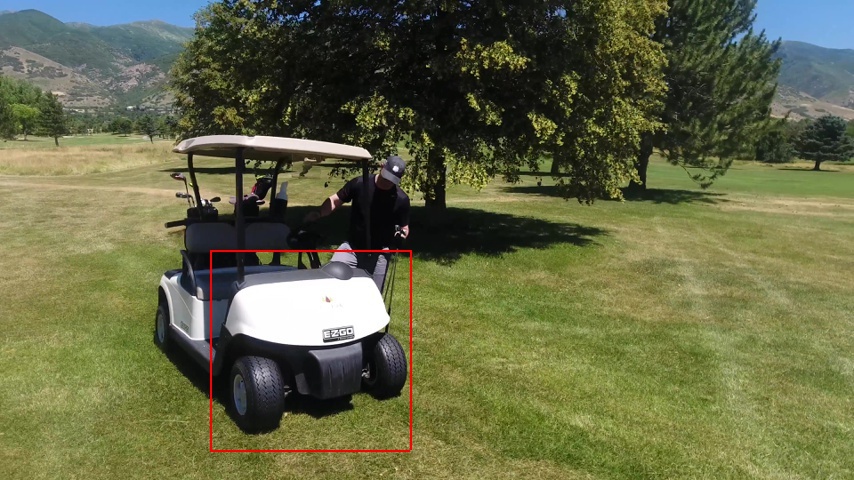}
		\caption{Original}
		\label{fig:golf_f17_s20_c_rect}
		\vspace*{6pt}
	\end{subfigure}
	\begin{subfigure}{0.32\textwidth}
	    \captionsetup{justification=centering}
		\includegraphics[width=\textwidth]{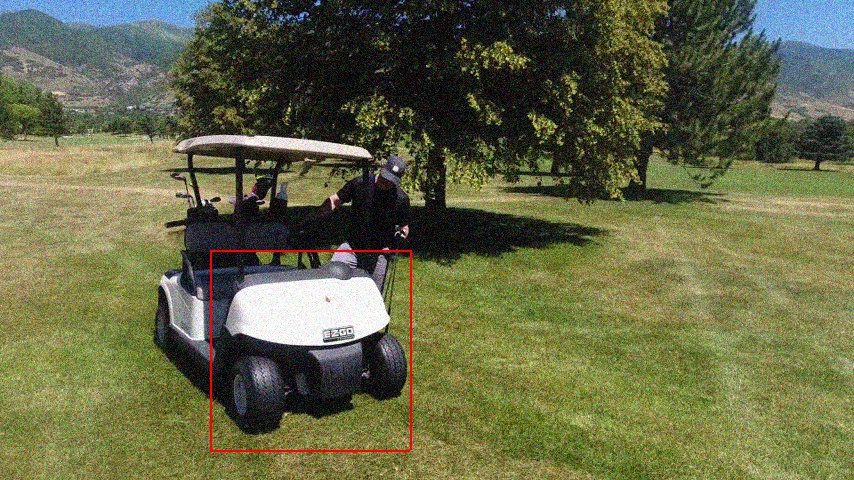}
		\caption{Noisy with $\sigma = 20$}
		\label{fig:golf_f17_s20_n_rect}
		\vspace*{6pt}
	\end{subfigure}
	\begin{subfigure}{0.32\textwidth}
	    \captionsetup{justification=centering}
		\includegraphics[width=\textwidth]{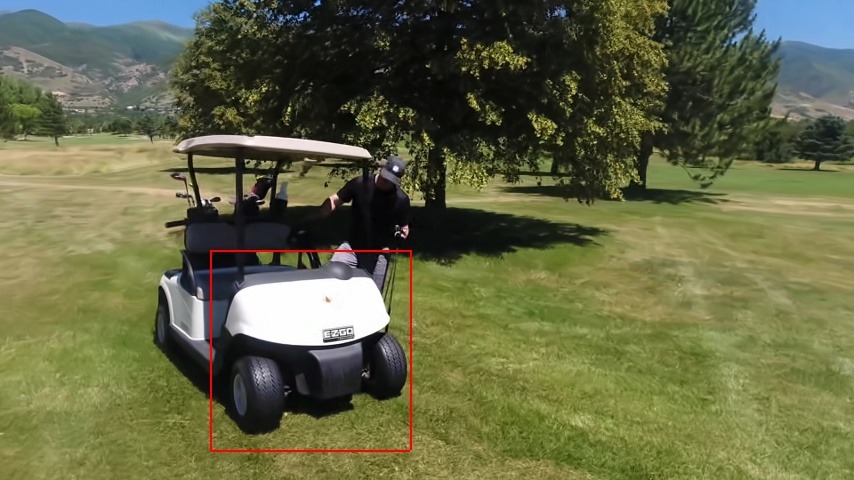}
		\caption{VNLB~\cite{arias2018video}, PSNR = 31.22dB}
		\label{fig:golf_f17_s20_vnlb_rect}
		\vspace*{6pt}
	\end{subfigure}
	\begin{subfigure}{0.32\textwidth}
	    \captionsetup{justification=centering}
		\includegraphics[width=\textwidth]{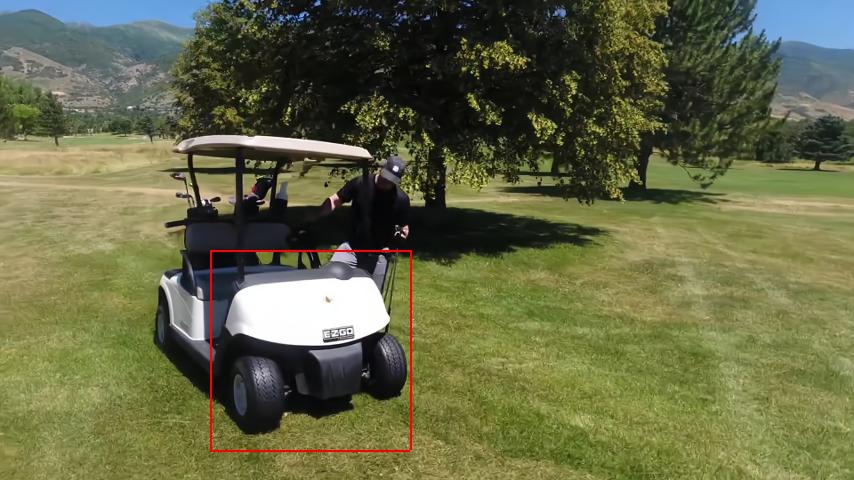}
		\caption{VNLnet~\cite{davy2019non}, PSNR = 30.19dB}
		\label{fig:golf_f17_s20_vnlnet_rect}
		\vspace*{6pt}
	\end{subfigure}
	\begin{subfigure}{0.32\textwidth}
	    \captionsetup{justification=centering}
		\includegraphics[width=\textwidth]{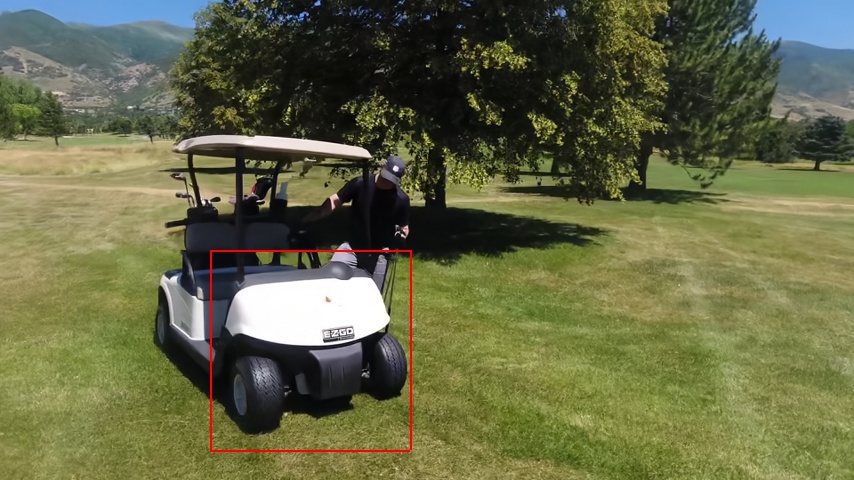}
		\caption{FastDVDnet~\cite{tassano2020fastdvdnet}, PSNR = 31.29dB}
		\label{fig:golf_f17_s20_fastdvd_rect}
		\vspace*{6pt}
	\end{subfigure}
	\begin{subfigure}{0.32\textwidth}
	    \captionsetup{justification=centering}
		\includegraphics[width=\textwidth]{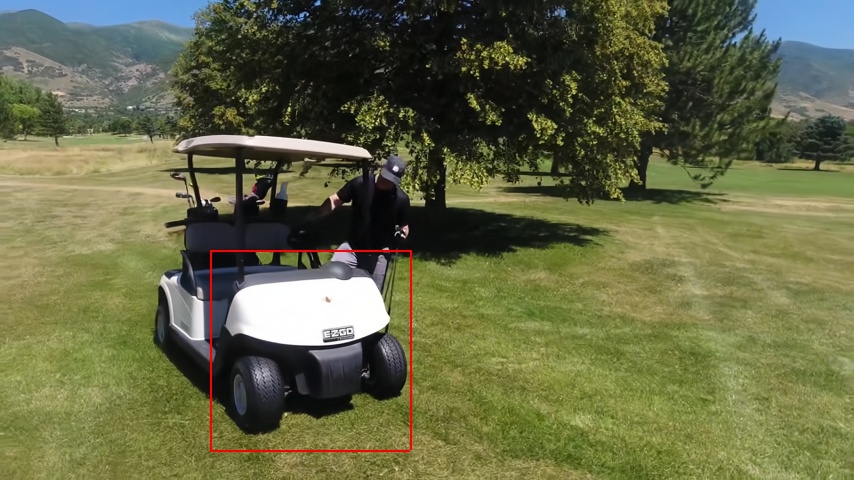}
		\caption{PaCNet (ours), PSNR = 32.14dB}
		\label{fig:golf_f17_s20_covid_rect}
		\vspace*{6pt}
	\end{subfigure}
	\begin{subfigure}{0.32\textwidth}
	    \captionsetup{justification=centering}
		\includegraphics[width=\textwidth]{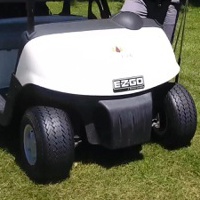}
		\caption{Original}
		\label{fig:golf_f17_s20_c_zoom}
		\vspace*{6pt}
	\end{subfigure}
	\begin{subfigure}{0.32\textwidth}
	    \captionsetup{justification=centering}
		\includegraphics[width=\textwidth]{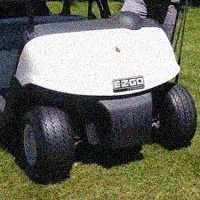}
		\caption{Noisy with $\sigma = 20$}
		\label{fig:golf_f17_s20_n_zoom}
		\vspace*{6pt}
	\end{subfigure}
	\begin{subfigure}{0.32\textwidth}
	    \captionsetup{justification=centering}
		\includegraphics[width=\textwidth]{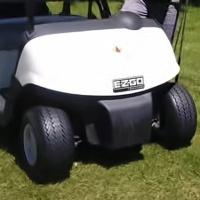}
		\caption{VNLB~\cite{arias2018video}, PSNR = 32.75dB}
		\label{fig:golf_f17_s20_vnlb_zoom}
		\vspace*{6pt}
	\end{subfigure}
	\begin{subfigure}{0.32\textwidth}
	    \captionsetup{justification=centering}
		\includegraphics[width=\textwidth]{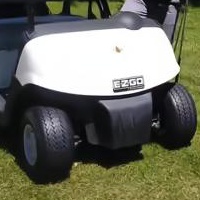}
		\caption{VNLnet~\cite{davy2019non}, PSNR = 31.15B}
		\label{fig:golf_f17_s20_vnlnet_zoom}
		\vspace*{6pt}
	\end{subfigure}
	\begin{subfigure}{0.32\textwidth}
	    \captionsetup{justification=centering}
		\includegraphics[width=\textwidth]{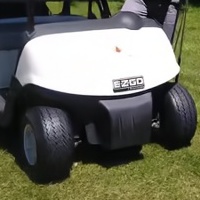}
		\caption{FastDVDnet~\cite{tassano2020fastdvdnet}, PSNR = 32.40dB}
		\label{fig:golf_f17_s20_fastdvd_zoom}
		\vspace*{6pt}
	\end{subfigure}
	\begin{subfigure}{0.32\textwidth}
	    \captionsetup{justification=centering}
		\includegraphics[width=\textwidth]{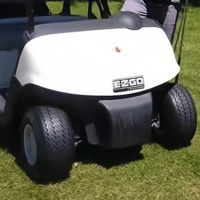}
		\caption{PaCNet (ours), PSNR = 33.42dB}
		\label{fig:golf_f17_s20_covid_zoom}
		\vspace*{6pt}
	\end{subfigure}
	\caption{Denoising example with $\sigma = 20$. The figure shows frame 18 of the sequence \emph{golf}. The PSNR values appearing in~\ref{fig:golf_f17_s20_vnlb_rect},~\ref{fig:golf_f17_s20_vnlnet_rect},~\ref{fig:golf_f17_s20_fastdvd_rect}~and~\ref{fig:golf_f17_s20_covid_rect} refer to the whole frame, whereas those in~\ref{fig:golf_f17_s20_vnlb_zoom},~\ref{fig:golf_f17_s20_vnlnet_zoom},~\ref{fig:golf_f17_s20_fastdvd_zoom}~and~\ref{fig:golf_f17_s20_covid_zoom} refer to the cropped area. As can be seen, PaCNet leads to better reconstructed results -- see the pattern on wheels.}
	\label{fig:golf_f17_s20}
\end{figure*}

\begin{table*}
    \centering
    \setlength{\doublerulesep}{1pt}
    \begin{tabular}{|c||c|c|c|c|c||c|}
        \hline
        \multirow{2}{*}{Method} & \multicolumn{5}{c||}{Noise $\sigma$} & \multirow{2}{*}{Average} \\
        \hhline{|~||-----||~}
        & 10 & 20 & 30 & 40 & 50 & \\
        \hhline{|=#=|=|=|=|=#=|}
        $\operatorname{DVDnet}$~\cite{tassano2019dvdnet} & 36.08 & 33.49 & 31.79 & 30.55 & 29.56 & 32.29 \\
        \hline
        $\operatorname{FastDVDnet}$~\cite{tassano2020fastdvdnet} & 36.44 & 33.43 & 31.68 & 30.46 & 29.53 & 32.31 \\
        \hline
        $\operatorname{PaCNet}$ (ours) & 37.06 & 33.94 & 32.05 & 30.70 & 29.66 & 32.68 \\
        \hline
    \end{tabular}
    \caption{Video denoising performance on Set8~\cite{tassano2019dvdnet}.}
    \label{tab:denoising_set8}
\end{table*}
Table~\ref{tab:denoising_set8} reports the average PSNR performance per noise level for 8 video sequences from the Set8~\cite{tassano2019dvdnet} dataset. Figure~\ref{fig:fr_psnr} presents graphs showing PSNR versus frame number for several test video sequences comparing PaCNet performance with VNLB~\cite{arias2018video}, VNLnet~\cite{davy2019non}, and FasDVDnet~\cite{tassano2020fastdvdnet}. Figures~\ref{fig:salsa_f8_s40},~\ref{fig:tractor_f22_s20}~and~\ref{fig:golf_f17_s20} show visual comparisons of our method versus leading algorithms. In addition to these figures, we attach to our paper several video (AVI) files that show comparisons of video sequences. Each file simultaneously plays the outcomes of four denoising algorithms: VNLB~\cite{arias2018video}, VNLnet~\cite{davy2019non}, FastDVDnet~\cite{tassano2020fastdvdnet}, and PaCNet (ours), along with the clean and the noisy sequences. These sequences are arranged according to the chart shown in Figure~\ref{fig:video_chart}.

Files \emph{salsa\_s40\_merge\_rect.avi} and \emph{skate-jump\_s20\_merge\_rect.avi} show the video sequences \emph{salsa} and \emph{skate-jump} contaminated by noise with ${\sigma = 40}$ and ${\sigma = 20}$ respectively. There are two rectangles, red and green, in each video. The rest four files show zoom-in on the area in these rectangles:
\begin{itemize}
    \item The green rectangle in \emph{salsa} is shown in \emph{salsa\_s40\_merge\_zoom\_g.avi}. As can be seen, our result is sharper than the VNLB outcome and less noisy than the outputs of FastDVDnet and VNLnet -- see for example the floor. Also, observe that the VNLnet  has noticeable artifacts around the legs.
    \item The red rectangle in \emph{salsa} is shown in \emph{salsa\_s40\_merge\_zoom\_r.avi}. As can be seen, PaCNet leads to better reconstruction -- see for example the brick wall. Our output is sharper and less noisy than the competitors' results.
    \item The green and the red rectangles of \emph{skate-jump} are shown in \emph{skate-jump\_s20\_merge\_zoom\_g.avi} and \emph{skate-jump\_s20\_merge\_zoom\_r.avi} respectively. As can be seen here as well, our algorithm leads to better reconstruction -- e.g. see the trees.
\end{itemize}
The videos are better seen in repeat mode.

\end{appendices}

\end{document}